# The LAMA Planner:
# Guiding Cost-Based Anytime Planning with Landmarks

**Silvia Richter**                                                                                          SILVIA.RICHTER@NICTA.COM.AU
*IIIS, Griffith University, Australia*
*and NICTA QRL, Australia*

**Matthias Westphal**                                                          WESTPHAM@INFORMATIK.UNI-FREIBURG.DE
*Albert-Ludwigs-Universität Freiburg*
*Institut für Informatik*
*Freiburg, Germany*

## Abstract

LAMA is a classical planning system based on heuristic forward search. Its core feature is the use of a pseudo-heuristic derived from *landmarks*, propositional formulas that must be true in every solution of a planning task. LAMA builds on the Fast Downward planning system, using finite-domain rather than binary state variables and multi-heuristic search. The latter is employed to combine the landmark heuristic with a variant of the well-known FF heuristic. Both heuristics are *cost-sensitive*, focusing on high-quality solutions in the case where actions have non-uniform cost. A weighted A* search is used with iteratively decreasing weights, so that the planner continues to search for plans of better quality until the search is terminated.

LAMA showed best performance among all planners in the sequential satisficing track of the International Planning Competition 2008. In this paper we present the system in detail and investigate which features of LAMA are crucial for its performance. We present individual results for some of the domains used at the competition, demonstrating good and bad cases for the techniques implemented in LAMA. Overall, we find that using landmarks improves performance, whereas the incorporation of action costs into the heuristic estimators proves not to be beneficial. We show that in some domains a search that ignores cost solves far more problems, raising the question of how to deal with action costs more effectively in the future. The iterated weighted A* search greatly improves results, and shows synergy effects with the use of landmarks.

## 1. Introduction

In the last decade, heuristic search has become the dominant approach to domain-independent satisficing planning. Starting with the additive heuristic by Bonet and Geffner (2001), implemented in the HSP planning system, much research has been conducted in search of heuristic estimators that are efficient to calculate yet powerful in guiding the search towards a goal state. The FF planning system by Hoffmann and Nebel (2001), using a heuristic estimator based on relaxed planning graphs, broke ground by showing best performance among all fully automated systems at the International Planning Competition in 2000, and continues to be state of the art today. Ever since, heuristic-search approaches have played a prominent role in the classical or sequential satisficing tracks of the biennial competition, with Fast Downward (Helmert, 2006) winning in 2004 and SG-Plan (Chen, Wah, & Hsu, 2006) placing first in 2006.

The LAMA planning system is the youngest member in this line, winning the sequential satisficing track at the International Planning Competition (IPC) in 2008. LAMA is a classical planning system based on heuristic search. It follows in the footsteps of HSP, FF, and Fast Downward and uses their earlier work in many respects. In particular, it builds on Fast Downward by extending it in three major ways:

1. **Landmarks.** In LAMA, Fast Downward's causal graph heuristic is replaced with a variant of the FF heuristic (Hoffmann & Nebel, 2001) and heuristic estimates derived from *landmarks*. Landmarks are propositional formulas that have to become true at some point in every plan for the task at hand (Porteous, Sebastia, & Hoffmann, 2001). LAMA uses landmarks to direct search towards those states where many landmarks have already been achieved. Via preferred operators, landmarks are also used as an additional source of search control which complements the heuristic estimates. In recent work, we have shown this use of landmarks in addition to the FF heuristic to improve performance, by leading to more problems being solved and shorter solution paths (Richter, Helmert, & Westphal, 2008).

2. **Action costs.** Both the landmark heuristic we proposed earlier (Richter et al., 2008) and the FF heuristic have been adapted to use action costs. However, LAMA does not focus purely on the *cost-to-go*, i.e., the estimated cost of reaching the goal from a given search node. There is a danger that a cost-sensitive planner may concentrate too much on finding a cheap plan, at the expense of finding a plan at all within a given time limit. LAMA weighs the estimated cost-to-go (as a measure of plan quality) against the estimated goal distance (as a measure of remaining search effort) by combining the values for the two estimates.

3. **Anytime search.** LAMA continues to search for better solutions until it has exhausted the search space or is interrupted. After finding an initial solution with a greedy best-first search, it conducts a series of weighted A$^*$ searches with decreasing weights, restarting the search each time from the initial state when an improved solution is found. In recent work, we have shown this approach to be very efficient on planning benchmarks compared to other anytime methods (Richter, Thayer, & Ruml, 2010).

At the International Planning Competition 2008, LAMA outperformed its competitors by a substantial margin. This result was not expected by its authors, as their previous work concerning LAMA's putative core feature, the landmark heuristic (Richter et al., 2008), showed some, but not tremendous improvement over the base configuration without landmarks. This paper aims to provide a reference description of LAMA as well as an extensive evaluation of its performance in the competition.

- **Detailed description of LAMA.** We present all distinguishing components of the planner in detail, describing how landmarks are generated and used in LAMA, how action costs are incorporated into the heuristic estimators and how the anytime search proceeds. Some aspects of LAMA have been presented in previous publications (Richter et al., 2008, 2010; Helmert, 2006). However, aspects that have not been adequately covered in those publications, in particular the procedure for finding landmarks, are described here in detail. Other relevant aspects described in previous work, like the landmark heuristic, are summarised for the convenience of the reader. Our aim is that this paper, together with previous ones, form a comprehensive picture of the LAMA system.

- **Experimental evaluation of LAMA.** Building on this, we conduct an experimental evaluation focusing on the aspects that differentiate LAMA from predecessor systems like FF and Fast Downward. We do not repeat comparisons published in earlier work, like the comparison between LAMA's anytime method and other anytime algorithms (Richter et al., 2010), or the comparison of LAMA's methods for handling landmarks to alternative landmark approaches (Richter et al., 2008). Instead, we aim to elicit how much the performance of the LAMA system as a whole is enhanced by each of the three distinguishing features described above (landmarks, action costs and anytime search). To answer this question, we contrast several variations of our planner using various subsets of these features.

We find that using cost-sensitive heuristics did *not* pay off on the IPC 2008 benchmark tasks. Our results show that the cost-sensitive variant of the FF heuristic used in LAMA performs significantly worse than the traditional unit-cost version of the same heuristic. Similarly, all other cost-sensitive planners in the competition fared worse than the baseline planner FF that ignored action costs, demonstrating that cost-based planning presents a considerable challenge. While we do not conduct a full analysis of the reasons for this, we showcase the problems of the cost-sensitive FF heuristic in some example domains and provide informed hypotheses for the encountered effects. Landmarks prove to be particularly helpful in this context. While in the unit-cost case landmarks only lead to a moderate increase in performance, in the case of planning with action costs they substantially improve coverage (the number of problems solved), thus effectively mitigating the problems of the cost-sensitive FF heuristic in LAMA. The anytime search significantly improves the quality of solutions throughout and even acts in synergy with landmarks in one domain.

## 2. Preliminaries

We use a planning formalism with state variables of finite (rather than binary) range, similar to the one employed by Helmert (2009). It is based on the SAS$^+$ planning model (Bäckström & Nebel, 1995), but extends it with conditional effects. While LAMA also handles axioms in the same way as Fast Downward (Helmert, 2006), we do not formalise axioms here, since they are not important for our purposes.

**Definition 1.** *Planning tasks in finite-domain representation (FDR tasks)*
*A **planning task in finite-domain representation (FDR task)** is given by a 5-tuple $\langle \mathcal{V}, s_0, s_\star, O, C \rangle$ with the following components:*

- $\mathcal{V}$ *is a finite set of **state variables**, each with an associated finite domain $\mathcal{D}_v$.*

  *A **fact** is a pair $\langle v, d \rangle$ (also written $v \mapsto d$), where $v \in \mathcal{V}$ and $d \in \mathcal{D}_v$. A **partial variable assignment** $s$ is a set of facts, each with a different variable. (We use set notation such as $\langle v, d \rangle \in s$ and function notation such as $s(v) = d$ interchangeably.) A **state** is a variable assignment defined on all variables $\mathcal{V}$.*

- $s_0$ *is a state called the **initial state**.*

- $s_\star$ *is a partial variable assignment called the **goal**.*

- $O$ *is a finite set of **operators**. An operator $\langle pre, eff \rangle$ consists of a partial variable assignment pre called its **precondition**, and a finite set of **effects** eff. Effects are triplets $\langle cond, v, d \rangle$, where cond is a (possibly empty) partial variable assignment called the **effect condition**, v is the affected variable and $d \in \mathcal{D}_v$ is called the **new value** for v.*

- $C: O \to \mathbb{N}_0^+$ *is an integer-valued non-negative* **action cost** *function.*

*An operator $o = \langle pre, eff \rangle \in O$ is* **applicable** *in a state $s$ if $pre \subseteq s$, and its effects are consistent, i. e., there is a state $s'$ such that $s'(v) = d$ for all $\langle cond, v, d \rangle \in eff$ where $cond \subseteq s$, and $s'(v) = s(v)$ otherwise. In this case, we say that the operator $o$ can be* **applied** *to $s$ resulting in the state $s'$ and write $s[o]$ for $s'$.*

*For operator sequences $\pi = \langle o_1, \ldots, o_n \rangle$, we write $s[\pi]$ for $s[o_1] \ldots [o_n]$ (only defined if each operator is applicable in the respective state). The operator sequence $\pi$ is called a* **plan** *if $s_\star \subseteq s_0[\pi]$. The* **cost** *of $\pi$ is the sum of the action costs of its operators, $\sum_{i=1}^n C(o_i)$.*

Each state variable $v$ of a planning task in finite-domain representation has an associated directed graph called the domain transition graph, which captures the ways in which the value of $v$ may change (Jonsson & Bäckström, 1998; Helmert, 2006). The vertex set of this graph is $\mathcal{D}_v$, and it contains an arc between two nodes $d$ and $d'$ if there exists an operator that can change the value of $v$ from $d$ to $d'$. Formally:

**Definition 2.** *Domain transition graph*
*The* **domain transition graph (DTG)** *of a state variable $v \in \mathcal{V}$ of an FDR task $\langle \mathcal{V}, s_0, s_\star, O, C \rangle$ is the digraph $\langle \mathcal{D}_v, A \rangle$ which includes an arc $\langle d, d' \rangle$ iff $d \neq d'$, there is an operator $\langle pre, eff \rangle \in O$ with $\langle cond, v, d' \rangle \in eff$, and for the union of conditions $pre \cup cond$ it holds that either it contains $v = d$ or it does not contain $v = \tilde{d}$ for any $\tilde{d} \in \mathcal{D}_v$.*

## 3. System Architecture

LAMA builds on the Fast Downward system (Helmert, 2006), inheriting the overall structure and large parts of the functionality from that planner. Like Fast Downward, LAMA accepts input in the PDDL2.2 Level 1 format (Fox & Long, 2003; Edelkamp & Hoffmann, 2004), including ADL conditions and effects and derived predicates (axioms). Furthermore, LAMA has been extended to handle the action costs introduced for IPC 2008 (Helmert, Do, & Refanidis, 2008). Like Fast Downward, LAMA consists of three separate components:

- The translation module

- The knowledge compilation module

- The search module

These components are implemented as separate programs that are invoked in sequence. In the following, we provide a brief description of the translation and knowledge compilation modules. The main changes in LAMA, compared to Fast Downward, are implemented in the search module, which we discuss in detail.

### 3.1 Translation

The *translation module*, short *translator*, transforms the PDDL input into a planning task in finite-domain representation as specified in Definition 1. The main components of the translator are an efficient grounding algorithm for instantiating schematic operators and axioms, and an invariant synthesis algorithm for determining groups of mutually exclusive facts. Such fact groups are consequently replaced by a single state variable, encoding *which* fact (if any) from the group is satisfied in a given world state. Details on this component can be found in a recent article by Helmert (2009).

The groups of mutually exclusive facts (mutexes) found during translation are later used to determine orderings between landmarks. For this reason, LAMA does not use the finite-domain representations offered at IPC 2008 (object fluents), but instead performs its own translation from binary to finite-domain variables. While not all mutexes computed by the translation module are needed for the new encoding of the planning task, the module has been extended in LAMA to retain all found mutexes for their later use with landmarks.

Further changes we made, compared to the translation module described by Helmert, were to add the capability of handling action costs, implement an extension concerning the parsing of complex operator effect formulas, and limit the runtime of the invariant synthesis algorithm. As invariant synthesis may be time critical, in particular on large (grounded) PDDL input, we limit the maximum number of considered mutex candidates in the algorithm, and abort it, if necessary, after five minutes. Note that finding few or no mutexes does not change the way the translation module works; if no mutexes are found, the resulting encoding of the planning task contains simply the same (binary-domain) state variables as the PDDL input. When analysing the competition results, we found that the synthesis algorithm had aborted only in some of the tasks of one domain (Cyber Security).

### 3.2 Knowledge Compilation

Using the finite-domain representation generated by the translator, the *knowledge compilation* module is responsible for building a number of data structures that play a central role in the subsequent landmark generation and search. Firstly, *domain transition graphs* (see Definition 2) are produced which encode the ways in which each state variable may change its value through operator applications and axioms. Furthermore, data structures are constructed for efficiently determining the set of applicable operators in a state and for evaluating the values of derived state variables. We refer to Helmert (2006) for more detail on the knowledge compilation component, which LAMA inherits unchanged from Fast Downward.

### 3.3 Search

The search module is responsible for the actual planning. Two algorithms for heuristic search are implemented in LAMA: (a) a greedy best-first search, aimed at finding a solution as quickly as possible, and (b) a weighted A* search that allows balancing speed against solution quality. Both algorithms are variations of the standard textbook methods, using open and closed lists. The greedy best-first search always expands a state with minimal heuristic value $h$ among all open states and never expands a state more than once. In order to encourage cost-efficient plans without incurring much overhead, it breaks ties between equally promising states by preferring those states that are reached by cheaper operators, i.e., taking into account the last operator on the path to the considered state in the search space. (The cost of the entire path could only be used at the expense of increased time or space requirements, so that we do not consider this.) Weighted A* search (Pohl, 1970) associates costs with states and expands a state with minimal $f'$-value, where $f' = w \cdot h + g$, the *weight w* is an integer $\geq 1$, and $g$ is the best known cost of reaching the considered state from the initial state. In contrast to the greedy search, weighted A* search re-expands states whenever it finds cheaper paths to them.

In addition, both search algorithms use three types of *search enhancements* inherited from Fast Downward (Helmert, 2006; Richter & Helmert, 2009). Firstly, multiple heuristics are employed within a *multi-queue* approach to guide the search. Secondly, *preferred operators* — similar to the helpful actions in FF — allow giving precedence to operators that are deemed more helpful than others in a state. Thirdly, *deferred heuristic evaluation* mitigates the impact of large branching factors assuming that heuristic estimates are fairly accurate. In the following, we discuss these techniques and the resulting algorithms in more detail and give pseudo code for the greedy best-first search. The weighted A* search is very similar, so we point out the differences between the two algorithms along the way.

**Multi-queue heuristic search.** LAMA uses two heuristic functions to guide its search: the name-giving landmark heuristic (see Section 5), and a variant of the well-known FF heuristic (see Section 6). The two heuristics are used with separate queues, thus exploiting strengths of the utilised heuristics in an orthogonal way (Helmert, 2006; Röger & Helmert, 2010). To this end, separate open lists are maintained for each of the two heuristics. States are always evaluated with respect to both heuristics, and their successors are added to all open lists (in each case with the value corresponding to the heuristic of that open list). When choosing which state to evaluate and expand next, the search algorithm alternates between the different queues based on numerical priorities assigned to each queue. These priorities are discussed later.

**Deferred heuristic evaluation.** The use of deferred heuristic evaluation means that states are not heuristically evaluated upon generation, but upon expansion, i. e., when states are generated in greedy best-first search, they are put into the open list not with their own heuristic value, but with that of their parent. Only after being removed from the open list are they evaluated heuristically, and their heuristic estimate is in turn used for their successors. The use of deferred evaluation in weighted A* search is analogous, using $f'$ instead of $h$ as the sorting criterion of the open lists. If many more states are generated than expanded, deferred evaluation leads to a substantial reduction in the number of heuristic estimates computed. However, deferred evaluation incurs a loss of heuristic accuracy, as the search can no longer use $h$-values or $f'$-values to differentiate between the successors of a state (all successors are associated with the parent's value in the open list). Preferred operators are very helpful in this context as they provide an alternative way to determine promising successors.

**Preferred operators.** Operators that are deemed particularly useful in a given state are marked as preferred. They are computed by the heuristic estimators along with the heuristic value of a state (see Sections 6 and 5). To use preferred operators, in the greedy best-first search as well as in the weighted A* search, the planner maintains an additional *preferred-operator queue* for each heuristic. When a state is evaluated and expanded, those successor states that are reached via a preferred operator (the *preferred states*) are put into the preferred-operator queues, in addition to being put into the regular queues like the non-preferred states. (Analogously to regular states, any state preferred by at least one heuristic is added to *all* preferred-operator queues. This allows for cross-fertilisation through information exchange between the different heuristics.) States in the preferred-operator queues are evaluated earlier on average, as they form part of more queues and have a higher chance of being selected at any point in time than the non-preferred states. In addition,

LAMA (like the IPC 2004 version of Fast Downward) gives even higher precedence to preferred successors via the following mechanism. The planner keeps a priority counter for each queue, initialised to 0. At each iteration, the next state is removed from the queue that has the highest priority. Whenever a state is removed from a queue, the priority of that queue is decreased by 1. If the priorities are not changed outside of this routine, this method will alternate between all queues, thus expanding states from preferred queues and regular queues equally often. To increase the use of preferred operators, LAMA increases the priorities of the preferred-operator queues by a large number *boost* of value 1000 whenever progress is made, i. e., whenever a state is discovered that has a better heuristic estimate than previously expanded states. Subsequently, the next 1000 states will be removed from preferred-operator queues. If another improving state is found within the 1000 states, the boosts accumulate and, accordingly, it takes longer until states from the regular queues are expanded again.

Alternative methods for using preferred operators include the one employed in the YAHSP system (Vidal, 2004), where preferred operators are always used over non-preferred ones. By contrast, our scheme does not necessarily empty the preferred queues before switching back to regular queues. In the FF planner (Hoffmann & Nebel, 2001), the emphasis on preferred operators is even stronger than in YAHPS: the search in FF is restricted to preferred operators until either a goal is found or the restricted search space has been exhausted (in which case a new search is started without preferred operators). Compared to these approaches, the method for using preferred operators in LAMA, in conjunction with deferred heuristic evaluation, has been shown to result in substantial performance improvement and deliver best results in the classical setting of operators with unit costs (Richter & Helmert, 2009). The choice of 1000 as the boost value is not critical here, as we found various values between 100 and 50000 to give similarly good results. Only outside this range does performance drop noticeably.

Note that when using action costs, the use of preferred operators may be even more helpful than in the classical setting. For example, if all operators have a cost of 0, a heuristic using pure cost estimates might assign the same heuristic value of 0 to all states in the state space, giving no guidance to search at all. Preferred operators, however, still provide the same heuristic guidance in this case as in the case with unit action costs. While this is an extreme example, similar cases appear in practice, e. g. in the IPC 2008 domain Openstacks, where all operators except for the one opening a new stack have an associated cost of 0.

**Pseudo code.** Algorithm 1 shows pseudo code for the greedy best-first search. The main loop (lines 25–36) runs until either a goal has been found (lines 27–29) or the search space has been exhausted (lines 32–33). The closed list contains all seen states and also keeps track of the links between states and their parents, so that a plan can be efficiently extracted once a goal state has been found (line 28). In each iteration of the loop, the search adds the current state (initially the start state) to the closed list and processes it (lines 30–31), unless the state has been processed before, in which case it is ignored (line 26). By contrast, weighted A* search processes states again whenever they are reached via a path with lower cost than before, and updates their parent links in the closed list accordingly. Then the search selects the next open list to be used (the one with highest priority, line 34), decreases its priority and extracts the next state to be processed (lines 35–36). The processing of a state includes calculating its heuristic values and preferred operators with both heuristics (lines 3–4), expanding it, and inserting the successors into the appropriate open

Global variables:

$\Pi = \langle \mathcal{V}, s_0, s_\star, O, C \rangle$ ▷ Planning task to solve

$reg_{FF}$, $pref_{FF}$, $reg_{LM}$, $pref_{LM}$ ▷ Regular and preferred open lists for each heuristic

$best\_seen\_value$ ▷ Best heuristic value seen so far for each heuristic

$priority$ ▷ Numerical priority for each queue

1: **function** EXPAND_STATE($s$)
2:      $progress \leftarrow$ False
3:      **for** $h \in \{FF, LM\}$ **do**
4:          $h(s)$, $preferred\_ops(h, s) \leftarrow$ heuristic value of $s$ and preferred operators given $h$
5:          **if** $h(s) < best\_seen\_value[h]$ **then**
6:              $progress \leftarrow$ True
7:              $best\_seen\_value[h] \leftarrow h(s)$
8:      **if** $progress$ **then** ▷ Boost preferred-operator queues
9:          $priority[pref_{FF}] \leftarrow priority[pref_{FF}] + 1000$
10:          $priority[pref_{LM}] \leftarrow priority[pref_{LM}] + 1000$
11:      $succesor\_states \leftarrow \{ s[o] \mid o \in O \text{ and } o \text{ applicable in } s \}$
12:      **for** $s' \in succesor\_states$ **do**
13:          **for** $h \in \{FF, LM\}$ **do**
14:              add $s'$ to queue $reg_h$ with value $h(s)$ ▷ Deferred evaluation
15:              **if** $s'$ reached by operator $o \in preferred\_ops(h, s)$ **then**
16:                  add $s'$ to queue $pref_{FF}$ with value $FF(s)$, and to queue $pref_{LM}$ with value $LM(s)$

17: **function** GREEDY_BFS_LAMA
18:      $closed\_list \leftarrow \emptyset$
19:      **for** $h \in \{FF, LM\}$ **do** ▷ Initialize FF and landmark heuristics
20:          $best\_seen\_value[h] \leftarrow \infty$
21:          **for** $l \in \{reg, pref\}$ **do** ▷ Regular and preferred open lists for each heuristic
22:              $l_h \leftarrow \emptyset$
23:              $priority[l_h] \leftarrow 0$
24:      $current\_state \leftarrow s_0$
25:      **loop**
26:          **if** $current\_state \notin closed\_list$ **then**
27:              **if** $s = s_\star$ **then**
28:                  extract plan $\pi$ by tracing current state back to initial state in closed list
29:                  **return** $\pi$
30:              $closed\_list \leftarrow closed\_list \cup \{current\_state\}$
31:              EXPAND_STATE($current\_state$)
32:          **if** all queues are empty **then**
33:              **return** failure ▷ No plan exists
34:          $q \leftarrow$ non-empty queue with highest priority
35:          $priority[q] \leftarrow priority[q] - 1$
36:          $current\_state \leftarrow$ POP_STATE($q$) ▷ Get lowest-valued state from queue $q$

Algorithm 1: The greedy best-first-search with search enhancements used in LAMA.

lists (lines 11–16). If it is determined that a new best state has been found (lines 5-7), the preferred-operator queues are boosted by 1000 (lines 8-10).

### 3.3.1 Restarting Anytime Search

LAMA was developed for the International Planning Competition 2008 and is tailored to the conditions of this competition in several ways. In detail, those conditions were as follows. While in previous competitions coverage, plan quality and runtime were all used to varying degrees in order to determine the effectiveness of a classical planning system, IPC 2008 introduced a new integrated performance criterion. Each operator in the PDDL input had an associated non-negative integer *action cost*, and the aim was to find a plan of lowest-possible total cost within a given time limit of 30 minutes per task. Given that a planner solves a task at all within the time limit, this new performance measure depends only on plan quality, not on runtime, and thus suggests guiding the search towards a cheapest goal rather than a closest goal as well as using all of the available time to find the best plan possible.

Guiding the search towards cheap goals may be achieved in two ways, both of which LAMA implements: firstly, the heuristics should estimate the *cost-to-go*, i. e., the cost of reaching the goal from a given state, rather than the *distance-to-go*, i. e., the number of operators required to reach the goal. Both the landmark heuristic and the FF heuristic employed in LAMA are therefore capable of using action costs. Secondly, the search algorithm should not only take the cost-to-go from a given state into account, but also the cost necessary for reaching that state. This is the case for weighted A* search as used in LAMA. To make the most of the available time, LAMA employs an *anytime* approach: it first runs a greedy best-first search, aimed at finding a solution as quickly as possible. Once a plan is found, it searches for progressively better solutions by running a series of weighted A* searches with decreasing weight. The cost of the best known solution is used for pruning the search, while decreasing the weight over time makes the search progressively less greedy, trading speed for solution quality.

Several anytime algorithms based on weighted A* have been proposed (Hansen & Zhou, 2007; Likhachev, Ferguson, Gordon, Stentz, & Thrun, 2008). Their underlying idea is to *continue* the weighted A* search past the first solution, possibly adjusting search parameters like the weight or pruning bound, and thus progressively find better solutions. The anytime approach used in LAMA differs from these existing algorithms in that we do *not* continue the weighted A* search once it finds a solution. Instead, we start a new weighted A* search, i. e., we discard the open lists of the previous search and re-start from the initial state. While resulting in some duplicate effort, these restarts can help overcome bad decisions made by the early (comparatively greedy) search iterations with high weight (Richter et al., 2010). This can be explained as follows: After finding a goal state $s_g$, the open lists will usually contain many states that are close to $s_g$ in the search space, because the ancestors of $s_g$ have been expanded; furthermore, those states are likely to have low heuristic values because of their proximity to $s_g$. Hence, if the search is continued (even after updating the open lists with lower weights), it is likely to expand most of the states around $s_g$ before considering states that are close to the initial state. This can be critical, as it means that the search is concentrating on improving the *end* of the current plan, as opposed to its beginning. A bad beginning of a plan, however, may have severe negative influence on its quality, as it may be impossible to improve the quality of the plan substantially without changing its early operators.

(a) initial search, $w = 2$

(b) continued search, $w = 1.5$

(c) restarted search, $w = 1.5$

Figure 1: The effect of low-$h$ bias. For all grid states generated by the search, $h$-values are shown above $f'$-values. (a) Initial weighted A* search finds a solution of cost 6. (b) Continued search expands many states around the previous Open list (grey cells), finding another sub-optimal solution of cost 6. (c) Restarted search quickly finds the optimal solution of cost 5.

Consider the example of a search problem shown in Figure 1. The task is to reach a goal state ($g1$ or $g2$) from the start state $s$ in a gridworld, where the agent can move with cost 1 to each of the 8 neighbours of a cell if they are not blocked. The heuristic values are inaccurate estimates of the straight-line goal distances of cells. In particular, the heuristic values underestimate distances in the left half of the grid. We conduct a weighted A* search with weight 2 in Figure 1a (assuming for simplicity a standard textbook search, i.e., no preferred operators and no deferred evaluation). Because the heuristic values to the left of $s$ happen to be lower than to the right of $s$, the search expands states to the left and finds goal $g1$ with cost 6. The grey cells are generated, but not expanded in this search phase, i.e., they are in the open list. In Figure 1b, the search continues with a reduced weight of 1.5. A solution with cost 5 consists in turning right from $s$ and going to $g2$. However, the search will first expand all states in the open list that have an $f'$-value smaller than 7. After expanding a substantial number of states, the second solution it finds is a path which starts off left of $s$ and takes the long way around the obstacle to $g2$, again with cost 6. If we instead *restart* with an empty open list after the first solution (Figure 1c), fewer states are expanded. The critical state to the right of $s$ is expanded quickly and the optimal path is found.

Note that in the above example, it is in particular the systematic errors of the heuristic values that leads the greedy search astray and makes restarts useful. In planning, especially when using deferred evaluation, heuristic values may also be fairly inaccurate, and restarts can be useful. In an experimental comparison on all tasks from IPC 1998 to IPC 2006 (Richter et al., 2010) this restarting approach performed notably better than all other tested methods, dominating similar algorithms based on weighted A* (Hansen, Zilberstein, & Danilchenko, 1997; Hansen & Zhou, 2007; Likhachev, Gordon, & Thrun, 2004; Likhachev et al., 2008), as well as other anytime approaches (Zhou & Hansen, 2005; Aine, Chakrabarti, & Kumar, 2007).

### 3.3.2 Using cost and distance estimates

Both heuristic estimators used in LAMA are cost-sensitive, aiming to guide the search towards high-quality solutions. Focusing a planner purely on action costs, however, may be dangerous, as cheap plans may be longer and more difficult to find, which in the worst case could mean that the planner fails to find a plan at all within the given time limit. Zero-cost operators present a particular challenge: since zero-cost operators can always be added to a search path "for free", even a cost-sensitive search algorithm like weighted A* may explore very long search paths without getting closer to a goal. Methods have been suggested that allow a trade-off between the putative cost-to-go and the estimated goal distance (Gerevini & Serina, 2002; Ruml & Do, 2007). However, they require the user to specify the relative importance of cost versus distance up-front, a choice that was not obvious in the context of IPC 2008. LAMA gives equal weight to the cost and distance estimates by adding the two values during the computation of its heuristic functions (for more details, see Sections 5 and 6). This measure is a very simple one, and its effect changes depending on the magnitude and variation of action costs in a problem: the smaller action costs are, the more this method favours short plans over cheap plans. For example, 5 zero-cost operators result in an estimated cost of 5, whereas 2 operators of cost 1 result in an estimated cost of 4. LAMA would thus prefer the 2 operators of cost 1 over the 5 zero-cost operators. By contrast, when the action costs in a planning task are larger than the length of typical plans, the cost estimates dominate the distance estimates and LAMA is completely guided by costs. Nevertheless this simple measure proves useful on the IPC 2008 benchmarks, outperforming pure cost search in our experiments. More so-phisticated methods for automatically balancing cost against distance (for example by normalising the action costs in a given task with respect to their mean or median) are a topic of future work.

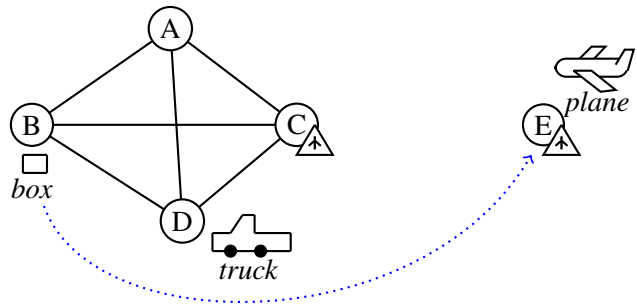

Figure 2: A simple Logistics task: transport the box from location *B* to location *E*.

## 4. Landmarks

Landmarks are subgoals that must be achieved in every plan. They were first introduced by Porteous, Sebastia and Hoffmann (2001) and were later studied in more depth by the same authors (Hoffmann, Porteous, & Sebastia, 2004). Using landmarks to guide the search for a solution in planning is an intuitive approach that humans might use. Consider the well-known benchmark domain Logistics, where the goal is to deliver objects (e. g. boxes) between various locations using a fleet of vehicles. Cities consist of sets of locations, where trucks may transport boxes within the city, whereas planes have to be used between cities. An example Logistics task is shown in Figure 2. Arguably the first mental step a human would perform, when trying to solve the task in Figure 2, is to realise that the box must be transported between the two cities, from the left city (locations A–D) to the right city (location E), and that therefore, *the box will have to be transported in the plane*. This in turn means that *the box will have to be at the airport location C*, so it can be loaded into a plane. This partitions the task into two subproblems, one of transporting the box to the airport at location *C*, and one of delivering it from there to the other city. Both subproblems are smaller and easier to solve than the original task.

Landmarks capture precisely these intermediate conditions that can be used to direct search: the facts $L_1$ = "*box* is at *C*" and $L_2$ = "*box* is in *plane*" are landmarks in the task shown in Figure 2. This knowledge, as well as the knowledge that $L_1$ must become true before $L_2$, can be automatically extracted from the task in a preprocessing step (Hoffmann et al., 2004).

LAMA uses landmarks to derive goal-distance estimates for a heuristic search. It measures the goal distance of a state by the number of landmarks that still need to be achieved on the path from this state to a goal. Orderings between landmarks are used to infer which landmarks should be achieved next, and whether certain landmarks have to be achieved more than once. In addition, *preferred operators* (Helmert, 2006) are used to suggest operators that achieve those landmarks that need to become true next. As we have recently shown, this method for using landmarks leads to substantially better performance than the previous use of landmarks by Hoffmann et al., both in terms of coverage and in terms of plan quality (Richter et al., 2008). We discuss the differences between their approach and ours in more detail in Section 4.3. In the following section we define landmarks and their orderings formally, including some useful special cases that can be detected efficiently.

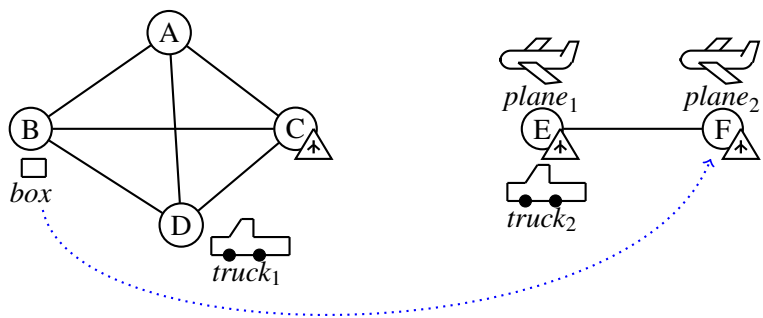

Figure 3: Extended logistics task: transport the box from location $B$ to location $F$.

## 4.1 Definitions

Hoffmann et al. (2004) define landmarks as facts that are true at some point in every plan for a given planning task. They also introduce *disjunctive landmarks*, defined as sets of facts of which at least one needs to be true at some point. We subsume their landmark definitions into a more general definition based on propositional formulas, as we believe this to be useful for future work on the topic of landmarks. It should be noted, however, that LAMA currently only supports fact landmarks and disjunctions of facts (for more details, see Section 4.2). Hoffmann et al. show that it is PSPACE-hard to determine whether a given fact is a landmark, and whether an ordering holds between two landmarks. Their complexity results carry over in a straight-forward way to the more general case of propositional formulas, so we do not repeat the proofs.

**Definition 3.** *Landmark*
*Let $\Pi = \langle \mathcal{V}, s_0, s_\star, \mathcal{O}, \mathcal{C} \rangle$ be a planning task in finite-domain representation, let $\pi = \langle o_1, \ldots, o_n \rangle$ be an operator sequence applicable in $s_0$, and let $i, j \in \{0, \ldots, n\}$.*

- *A propositional formula $\varphi$ over the facts of $\Pi$ is called a **fact formula**.*

- *A fact $F$ **is true at time** $i$ in $\pi$ iff $F \in s_0[\langle o_1, \ldots, o_i \rangle]$.*

- *A fact formula $\varphi$ **is true at time** $i$ in $\pi$ iff $\varphi$ holds given the truth value of all facts of $\Pi$ at time $i$. At any time $i < 0$, $\varphi$ is not considered true.*

- *A fact formula $\varphi$ is a **landmark** of $\Pi$ iff in each plan for $\Pi$, $\varphi$ is true at some time.*

- *A propositional formula $\varphi$ over the facts of $\Pi$ **is added at time** $i$ in $\pi$ iff $\varphi$ is true at time $i$ in $\pi$, but not at time $i - 1$ (it is considered added at time $0$ if it is true in $s_0$).*

- *A fact formula $\varphi$ **is first added at time** $i$ in $\pi$ iff $\varphi$ is true at time $i$ in $\pi$, but not at any time $j < i$.*

Note that facts in the initial state and facts in the goal are always landmarks by definition.

The landmarks we discussed earlier for the example task in Figure 2 were all facts. However, more complex landmarks may be required in larger tasks. Consider an extended version of the example, where the city on the right has two airports, and there are multiple planes and trucks, as depicted in Figure 3. The previous landmark $L_1$ = "*box* is at $C$" is still a landmark in our extended example. However, $L_2$ = "*box* is in *plane*" has no corresponding fact landmark in this task, since neither "*box* is in $plane_1$" nor "*box* is in $plane_2$" is a landmark. The disjunction "*box* is in $plane_1$ $\lor$ *box* is in $plane_2$", however, is a landmark. In the following we refer to landmarks that are facts as *fact landmarks*, and to disjunctions of facts as *disjunctive landmarks*. While the use of disjunctive landmarks has been shown to improve performance, compared to using only fact landmarks (Richter et al., 2008), more complex landmarks introduce additional difficulty both with regard to their detection and their handling during planning. As mentioned before, LAMA currently only uses fact landmarks and disjunctive landmarks, rather than general propositional formulas. The extension to more complex types of landmarks is an interesting topic of future work. (See Keyder, Richter and Helmert, 2010, for a discussion of *conjunctive landmarks*).

Various kinds of *orderings* between landmarks can be defined and exploited during the planning phase. We define three types of orderings for landmarks, which are equivalent formulations of the definitions by Hoffmann et al. (2004) adapted to the FDR setting:

**Definition 4. *Orderings between landmarks***
*Let $\varphi$ and $\psi$ be landmarks in an FDR planning task $\Pi$.*

- *We say that there is a **natural ordering** between $\varphi$ and $\psi$, written $\varphi \rightarrow \psi$, if in each plan where $\psi$ is true at time $i$, $\varphi$ is true at some time $j < i$.*

- *We say that there is a **necessary ordering** between $\varphi$ and $\psi$, written $\varphi \rightarrow_n \psi$, if in each plan where $\psi$ is added at time $i$, $\varphi$ is true at time $i - 1$.*

- *We say that there is a **greedy-necessary ordering** between $\varphi$ and $\psi$, written $\varphi \rightarrow_{gn} \psi$, if in each plan where $\psi$ is* first added *at time $i$, $\varphi$ is true at time $i - 1$.*

Natural orderings are the most general; every necessary or greedy-necessary ordering is natural, but not vice versa. Similarly, every necessary ordering is greedy-necessary, but not vice versa. Knowing that a natural ordering is also necessary or greedy-necessary allows deducing additional information about plausible temporal relationships between landmarks, as described later in this section. Also, the landmark heuristic in LAMA uses this knowledge to deduce whether a landmark needs to be achieved more than once. As a theoretical concept, necessary orderings ($\varphi$ is *always* true in the step before $\psi$) are more straightforward and appealing than greedy-necessary orderings ($\varphi$ is true in the step before $\psi$ becomes true *for the first time*). However, methods that find landmarks in conjunction with orderings can often find many more landmarks when using the more general concept of greedy-necessary orderings (Hoffmann et al., 2004). LAMA follows this paradigm and finds greedy-necessary (as well as natural) orderings, but not necessary orderings. In our example in Figure 3, "*box* is in $truck_1$" must be true before "*box* is at $C$" and also before "*box* is at $F$". The first of these orderings is greedy-necessary, but not necessary, and the second is neither greedy-necessary nor necessary, but natural.

Hoffmann et al. (2004) propose further kinds of orderings between landmarks that can be usefully exploited. For example, *reasonable orderings*, which were first introduced in the context of top-level goals (Koehler & Hoffmann, 2000), are orderings that do not necessarily hold in a given planning task. However, adhering to these orderings may save effort when solving the task. In our example task, it is "reasonable" to load the box onto $truck_1$ before driving the truck to the airport at $C$. However, this order is not guaranteed to hold in every plan, as it is *possible*, though not "reasonable", to drive the truck to $C$ first, then drive to $B$ and collect the box, and then return to $C$. The idea is that if a landmark $\psi$ must become false in order to achieve a landmark $\varphi$, but $\psi$ is needed after $\varphi$, then it is reasonable to achieve $\varphi$ before $\psi$ (as otherwise, we would have to achieve $\psi$ twice). The idea may be applied iteratively, as we are sometimes able to find new, induced reasonable orderings if we restrict our focus to plans that obey a first set of reasonable orderings. Hoffmann et al. call the reasonable orderings found in such a second pass *obedient-reasonable orderings*. The authors note that conducting more than two iterations of this process is not worthwhile, as it typically does not result in notable additional benefit. The following definition characterises these two types of orderings formally.

**Definition 5.** *Reasonable orderings between landmarks*
*Let $\varphi$ and $\psi$ be landmarks in an FDR planning task $\Pi$.*

- *We say that there is a **reasonable ordering** between $\varphi$ and $\psi$, written $\varphi \to_r \psi$, if for every plan $\pi$ where $\psi$ is added at time $i$ and $\varphi$ is first added at time $j$ with $i < j$, it holds that $\psi$ is not true at time $m$ with $m \in \{i + 1, \dots, j\}$ and $\psi$ is true at some time $k$ with $j \le k$.*

- *We say that a plan $\pi$ **obeys** a set of orderings $O$, if for all orderings $\varphi \to_x \psi \in O$, regardless of their type, it holds that $\varphi$ is first added at time $i$ in $\pi$ and $\psi$ is not true at any time $j \le i$.*

- *We say that there is an **obedient-reasonable ordering** between $\varphi$ and $\psi$ with regard to a set of orderings $O$, written $\varphi \to_r^O \psi$, if for every plan $\pi$ obeying $O$ where $\psi$ is added at time $i$ and $\varphi$ is first added at time $j$ with $i < j$, it holds that $\psi$ is not true at time $m$ with $m \in \{i + 1, \dots, j\}$ and $\psi$ is true at some time $k$ with $j \le k$.*

Our definitions are equivalent to those of Hoffmann et al. (2004), except that we care only about plans rather than arbitrary operator sequences, allowing us to (theoretically) identify more reasonable orderings. In practice, we use the same approximation techniques as Hoffmann et al., thus generating the same orderings.

A problem with reasonable and obedient-reasonable orderings is that they may be cyclic, i.e., chains of orderings $\varphi \to_r \psi \to_x \dots \to_r \varphi$ for landmarks $\varphi$ and $\psi$ may exist (Hoffmann et al., 2004). This is not the case for natural orderings, as their definition implies that they cannot be cyclic in solvable tasks.

In addition, the definitions as given above are problematic in special cases. Note that the definition of a reasonable ordering $\varphi \to_r \psi$ includes the case where there exist no $i < j$ such that $\psi$ is added at time $i$ and $\varphi$ is first added at time $j$, i.e., the case where it holds that in all plans $\varphi$ is first added (a) before or (b) at the same time as $\psi$.[1] While (a) implies that reasonable orderings are a generalisation of natural orderings, which might be regarded as a desirable property, (b) may lead to undesirable orderings. For example, it holds that $\varphi \to_r \psi$ and $\psi \to_r \varphi$ for all pairs $\varphi, \psi$ that are first added at the same time in all plans, for instance if $\varphi$ and $\psi$ are both true in the initial state. Similarly, it holds that $\varphi \to_r \varphi$ for all $\varphi$. We use these definitions despite their weaknesses here, and simply note that our planner does not create all such contentious orderings. LAMA does not create reflexive orderings $\varphi \to_r \varphi$; and $\varphi \to_r \psi$ with $\varphi, \psi$ true in the initial state is only created if it is assumed or proven that $\psi$ must be true strictly after $\phi$ at some point in any plan (see also Section 4.2.5). A re-definition of reasonable orderings, addressing the problems of the definition by Hoffmann et al. and identifying precisely the wanted/unwanted cases, is a topic of future work. Closely connected is the question whether reasonable orderings should be interpreted as strict orderings, where $\varphi$ should be achieved before $\psi$ (as in the definition of obedience above), or whether we allow achieving $\varphi$ and $\psi$ simultaneously. We use the strict sense of obedience for reasons of consistency with the previous work by Hoffmann et al., and because it aligns better with our intended meaning of reasonable orderings, even though this strict interpretation of obedience does not fit the contentious cases discussed above.

---

1. According to personal communication with the authors, this case was overlooked by Hoffmann et al.

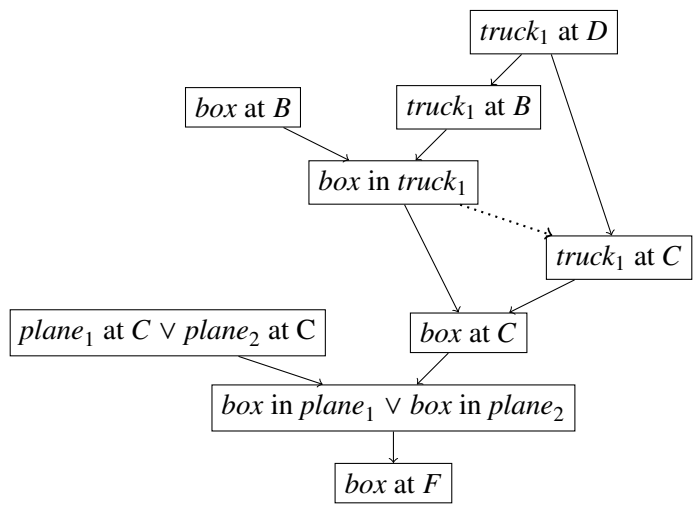

Figure 4: Partial landmark graph for the example task shown in Figure 3. Bold arcs represent natural orderings, dashed arcs represent reasonable orderings.

Landmarks and orderings may be represented using a directed graph called the *landmark graph*. A partial landmark graph for our extended example is depicted in Figure 4. The following section 4.2 contains an extensive description of how landmarks and their orderings are discovered in LAMA. Readers not interested in the exact details of this process may skip this description, as it is not central to the rest of this paper. Section 4.3 discusses how our approach for finding and using landmarks relates to previous work. Section 5 describes how landmarks are used as a heuristic estimator in LAMA.

## 4.2 Extracting Landmarks and Orderings

As mentioned before, deciding whether a given formula is a landmark and deciding orderings between landmarks are PSPACE-hard problems. Thus, practical methods for finding landmarks are incomplete (they may fail to find a given landmark or ordering) or unsound (they may falsely declare a formula to be a landmark, or determine a false ordering). Several polynomial methods have been proposed for finding fact landmarks and disjunctive landmarks, such as back-chaining from the goals of the task, using criteria based on the relaxed planning graph (Porteous et al., 2001; Hoffmann et al., 2004; Porteous & Cresswell, 2002), and forward propagation in the planning graph (Zhu & Givan, 2003).

The algorithm used in LAMA for finding landmarks and orderings between them is partly based on the previous back-chaining methods mentioned above, adapting them to the finite-domain representation including conditional effects. In addition, our algorithm exploits the finite-domain representation by using domain transition graphs to find further landmarks. We discuss the differences between our method and the previous ones in detail in Section 4.3. The idea of back-chaining is to start from a set of known landmarks and to find new fact landmarks or disjunctive landmarks that must be true in any plan *before* an already known landmark may become true. This procedure starts from the set of all goal facts, and stops when no more new landmarks can be found. Our method identifies new landmarks and orderings by considering, for any given fact landmark or disjunctive landmark $\psi$ that is not true in the initial state:

- The *shared preconditions* of its *possible first achievers*. These are the operator preconditions and effect conditions shared by all effects that can potentially first achieve $\psi$. This method has been adapted from previous work (see Section 4.3).

- For fact landmarks $v \mapsto d$, the domain transition graph (DTG) of $v$. Here, we identify nodes in the DTG (i. e., values $d'$ of $v$) that must necessarily be traversed in order to reach $d$.

- A restricted relaxed planning graph lacking all operators that could possibly achieve $\psi$. (There are some subtleties involving conditional effects that will be explained later.) Every landmark which does not occur in the last level of this graph can only be achieved after $\psi$.

As in previous work (Porteous et al., 2001; Hoffmann et al., 2004), we subsequently use the discovered landmarks and orderings to derive reasonable and obedient-reasonable orderings in a post-processing step. In the following, we give a detailed description of each step of the procedure for finding landmarks and orderings in LAMA. High-level pseudo code for our algorithm, containing the steps described in the following sections 4.2.1–4.2.4, is shown in Algorithm 2.

### 4.2.1 Back-Chaining: Landmarks via Shared Preconditions of Possible First Achievers

*First achievers* of a fact landmark or disjunctive landmark $\psi$ are those operators that potentially make $\psi$ true and can be applied at the end of a partial plan that has never made $\psi$ true before. We call any fact $A$ that is a precondition for *each* of the first achievers a *shared precondition*. As at least one of the first achievers must be applied to make $\psi$ true, $A$ must be true before $\psi$ can be achieved, and $A$ is thus a landmark, with the ordering $A \rightarrow_{gn} \psi$. Any effect condition for $\psi$ in an operator can be treated like a precondition in this context, as we are interested in finding the conditions that must hold for $\psi$ to become true. We will in the following use the term *extended preconditions* of an operator $o$ for $\psi$ to denote the union of the preconditions of $o$ and its effect conditions for $\psi$. The extended preconditions shared by all achievers of a fact are calculated in line 19 of Algorithm 2. In addition, we can create disjunctive landmarks $\varphi$ by selecting, from the precondition facts of the first achievers, sets of facts such that each set contains one extended precondition fact from each first achiever (line 22). As one of the first achievers must be applied to make $\psi$ true, one of the facts in $\varphi$ must be true before $\psi$, and the disjunction $\varphi$ is thus a landmark, with the ordering $\varphi \rightarrow_{gn} \psi$. Since the number of such disjunctive landmarks is exponential in the number of achievers of $\psi$, we restrict ourselves to disjunctions where all facts stem from the same predicate symbol, which are deemed most helpful (Hoffmann et al., 2004). Furthermore, we discard any fact sets of size greater than four, though we found this restriction to have little impact compared to the predicate restriction.

Global variables:
  $\Pi = \langle \mathcal{V}, s_0, s_\star, O, C \rangle$        ▷ Planning task to solve
  $LG = \langle L, O \rangle$        ▷ Landmark graph of $\Pi$
  *queue*        ▷ Landmarks to be back-chained from

1: **function** ADD_LANDMARK_AND_ORDERING($\varphi, \varphi \rightarrow_x \psi$)
2:    **if** $\varphi$ is a fact and $\nexists \chi \in L: \chi \not\equiv \varphi$ and $\varphi \models \chi$ **then**     ▷ Prefer fact landmarks
3:      $L \leftarrow L \setminus \{\chi\}$        ▷ Remove disjunctive landmark
4:      $O \leftarrow O \setminus \{(\vartheta \rightarrow_x \chi), (\chi \rightarrow_x \vartheta) \mid \vartheta \in L\}$     ▷ Remove obsolete orderings
5:    **if** $\exists \chi \in L: \chi \not\equiv \varphi$ and $var(\varphi) \cap var(\chi) \neq \emptyset$ **then**   ▷ Abort on overlap with existing landmark
6:      **return**
7:    **if** $\varphi \notin L$ **then**        ▷ Add new landmark to graph
8:      $L \leftarrow L \cup \{\varphi\}$
9:      *queue* $\leftarrow$ *queue* $\cup \{\varphi\}$
10:    $O \leftarrow O \cup \{\varphi \rightarrow_x \psi\}$        ▷ Add new ordering to graph

11: **function** IDENTIFY_LANDMARKS
12:    $LG \leftarrow \langle s_\star, \emptyset \rangle$        ▷ Landmark graph starts with all goals, no orderings
13:    *queue* $\leftarrow s_\star$
14:    *further_orderings* $\leftarrow \emptyset$        ▷ Additional orderings (see Section 4.2.3)
15:    **while** *queue* $\neq \emptyset$ **do**
16:      $\psi \leftarrow$ POP(*queue*)
17:      **if** $s_0 \not\models \psi$ **then**
18:        *RRPG* $\leftarrow$ the restricted relaxed plan graph for $\psi$
19:        $pre_{shared} \leftarrow$ shared extended preconditions for $\psi$ extracted from *RRPG*
20:        **for** $\varphi \in pre_{shared}$ **do**
21:          ADD_LANDMARK_AND_ORDERING($\varphi, \varphi \rightarrow_{gn} \psi$)
22:        $pre_{disj} \leftarrow$ sets of facts covering shared extended preconditions for $\psi$ given *RRPG*
23:        **for** $\varphi \in pre_{disj}$ **do**
24:          **if** $s_0 \not\models \varphi$ **then**
25:            ADD_LANDMARK_AND_ORDERING($\varphi, \varphi \rightarrow_{gn} \psi$)
26:      **if** $\psi$ is a fact **then**
27:        $pre_{lookahead} \leftarrow$ extract landmarks from DTG of the variable in $\psi$ using RRPG
28:        **for** $\varphi \in pre_{lookahead}$ **do**
29:          ADD_LANDMARK_AND_ORDERING($\varphi, \varphi \rightarrow \psi$)
30:      *potential_orderings* $\leftarrow$ *potential_orderings* $\cup \{\psi \rightarrow F \mid F$ never true in RRPG $\}$
31:    add further orderings between landmarks from *potential_orderings*

Algorithm 2: Identifying landmarks and orderings via back-chaining, domain transition graphs and restricted relaxed planning graphs.

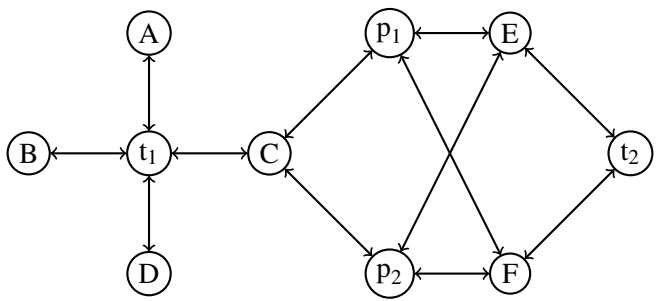

Figure 5: Domain transition graph for the location of the box in our extended example (Figure 3).

Since it is PSPACE-hard to determine the set of first achievers of a landmark $\psi$ (Hoffmann et al., 2004), we use an over-approximation containing every operator that can *possibly* be a first achiever (Porteous & Cresswell, 2002). By intersecting over the extended preconditions of (possibly) more operators we do not lose correctness, though we may miss out on some landmarks. The approximation of first achievers of $\psi$ is done with the help of a *restricted relaxed planning graph*. During construction of the graph we leave out any operators that would add $\psi$ unconditionally, and we also ignore any conditional effects which could potentially add $\psi$. When the relaxed planning graph levels out, its last set of facts is an over-approximation of the facts that can be achieved *before $\psi$* in the planning task. Any operator that is applicable given this over-approximating set and achieves $\psi$ is a *possible first achiever* of $\psi$.

### 4.2.2 Landmarks via Domain Transition Graphs

Given a fact landmark $L = \{v \mapsto l\}$, we can use the domain transition graph of $v$ to find further fact landmarks $v \mapsto l'$ (line 27) as follows. If the DTG contains a node that occurs on *every* path from the *initial state value* $s_0(v)$ of a variable to the *landmark value* $l$, then that node corresponds to a landmark value $l'$ of $v$: We know that every plan achieving $L$ requires that $v$ takes on the value $l'$, hence the fact $L' = \{v \mapsto l'\}$ can be introduced as a new landmark and ordered naturally before $L$. To find these kinds of landmarks, we iteratively remove one node from the DTG and test with a simple graph algorithm whether $s_0(v)$ and $l$ are still connected – if not, the removed node corresponds to a landmark. We further improve this procedure by removing, as a preprocessing step, all nodes for which we know that they cannot be true before achieving $L$. These are the nodes that correspond to facts other than $L$ and do not appear in the restricted RPG that never adds $L$. Removing these nodes may decrease the number of paths reaching $L$ and may thus allow us to find more landmarks.

Consider again the landmark graph of our extended example, shown in Figure 4. Most of its landmarks and orderings can be found via the back-chaining procedure described in the previous section, because the landmarks are direct preconditions for achieving their successors in the graph. There are two exceptions: "*box* in *truck*$_1$" and "*box* at *C*". These two landmarks are however found with the DTG method. The DTG in Figure 5 immediately shows that the box location must take on both the value $t_1$ and the value $C$ on any path from its initial value $B$ to its goal value $F$.

### 4.2.3 ADDITIONAL ORDERINGS FROM RESTRICTED RELAXED PLANNING GRAPHS

The restricted relaxed planning graph (RRPG) described in Section 4.2.1, which for a given landmark $\psi$ leaves out all operators that could possibly achieve $\psi$, can be used to extract additional orderings between landmarks. Any landmark $\chi$ that does not appear in this graph cannot be reached before $\psi$, and we can thus introduce a natural ordering $\psi \rightarrow \chi$. For efficiency reasons, we construct the RRPG for $\psi$ only once (line 18), i.e., when needed to find possible first achievers of $\psi$ during the back-chaining procedure. We then extract all orderings between $\psi$ and facts that can only be reached after $\psi$ (line 30). For all such facts $F$ that are later recognised to be landmarks, we then introduce the ordering $\psi \rightarrow F$ (line 31).

### 4.2.4 OVERLAPPING LANDMARKS

Due to the iterative nature of the algorithm it is possible that we find disjunctive landmarks for which at least one of the facts is already known to be a fact landmark. In such cases, we let fact landmarks take precedence over disjunctive ones, i.e., when a disjunctive landmark is discovered that includes an already known fact landmark, we do not add the disjunctive landmark. Conversely, as soon as a fact landmark is found that is part of an already known disjunctive landmark, we discard the disjunctive landmark including its orderings[2], and add the fact landmark instead. To keep the procedure and the resulting landmark graph simple, we furthermore do not allow landmarks to overlap. Whenever some fact from a newly discovered disjunctive landmark is also part of some already known landmark, we do not add the newly discovered landmark. All these cases are handled in the function ADD-LANDMARK-AND-ORDERING (lines 1– 10).

### 4.2.5 GENERATING REASONABLE AND OBEDIENT-REASONABLE ORDERINGS

We want to introduce a reasonable ordering $L \rightarrow_r L'$ between two (distinct) fact landmarks $L$ and $L'$ if it holds that (a) $L'$ must be true at the same time or after first achieving $L$, and (b) achieving $L'$ before $L$ would require making $L'$ false again to achieve $L$. We approximate both (a) and (b) as proposed by Hoffmann et al. (2004) with sufficient conditions. In the case of (a), we test if $L' \in s_\star$ or if we have a chain of natural or greedy-necessary orderings between landmarks $L = L_1 \rightarrow \ldots \rightarrow L_n$, with $n > 1$, $L_{n-1} \neq L'$ and a greedy-necessary ordering $L' \rightarrow_{gn} L_n$. For (b) we check whether (i) $L$ and $L'$ are inconsistent, i.e., mutually exclusive, or (ii) all operators achieving $L$ have an effect that is inconsistent with $L'$, or (iii) there is a landmark $L''$ inconsistent with $L'$ with the ordering $L'' \rightarrow_{gn} L$.

Inconsistencies between facts can be easily identified in the finite-domain representation if the facts are of the form $v \mapsto d$ and $v \mapsto d'$, i.e., if they map the same variable to different values. In addition, LAMA uses the groups of inconsistent facts computed by its translator component.

In a second pass, obedient-reasonable orderings are added. This is done with the same method as above, except that now reasonable orderings are used in addition to natural and greedy-necessary orderings to derive the fact that a landmark $L'$ must be true after a landmark $L$. Finally, we use a simple greedy algorithm to break possible cycles due to reasonable and obedient-reasonable orderings in the landmark graph, where every time a cycle is identified, one of the involved reasonable or obedient-reasonable orderings is removed. The algorithm removes obedient-reasonable orderings rather than reasonable orderings whenever possible.

---

2. Note that an ordering $\{F, G\} \rightarrow \psi$ neither implies $F \rightarrow \psi$ nor $G \rightarrow \psi$ in general. Conversely, $\varphi \rightarrow \{F, G\}$ neither implies $\varphi \rightarrow F$ nor $\varphi \rightarrow G$.

### 4.3 Related Work

Orderings between landmarks are a generalisation of *goal orderings*, which have been frequently exploited in planning and search in the past. In particular, the approach by Irani and Cheng (Irani & Cheng, 1987; Cheng & Irani, 1989) is a preprocessing procedure like ours that analyses the planning task to extract necessary orderings between goals, which are then imposed on the search algorithm. A goal *A* is ordered before a goal *B* in this approach if in any plan *A* is necessarily true before *B*. Koehler and Hoffmann (2000) introduce *reasonable* orderings for goals.

Hoffmann et al. (2004), in an article detailing earlier work by Porteous et al. (2001), introduce the idea of landmarks, generalise necessary and reasonable orderings from goals to landmarks, and propose methods for finding and using landmarks for planning. The proposed method for finding landmarks, which was subsequently extended by Porteous and Cresswell (2002), is very closely related to ours. Hoffmann et al. propose a method for finding fact landmarks that proceeds in three stages. First, potential landmarks and orderings are suggested by a fast *candidate generation procedure*. Second, a *filtering procedure* evaluates a sufficient condition for landmarks on each candidate fact, removing those which fail the test. Third, reasonable and obedient-reasonable orderings between the landmarks are approximated. This step is largely identical in their approach and ours, except that we use different methods to recognise inconsistencies between facts.

The generation of landmark candidates is done via back-chaining from the goal much like in our approach, and intersecting preconditions over all operators which can first achieve a fact *F* and *appear before F in the relaxed planning graph*. Note that even if all these operators share a common precondition *L*, there might be other first achievers of *F* (appearing after *F* in the relaxed planning graph) that do not have *L* as a precondition, and hence *L* is not a landmark. To test whether a landmark candidate *L* found via back-chaining is indeed a landmark, Hoffmann et al. (2004) build a restricted relaxed planning task leaving out all operators which could add *L*. If this task is unsolvable, then *L* is a landmark. This is a sufficient, but not necessary condition: if *L* is necessary for solving the relaxed task it is also necessary for solving the original task, while the converse is not true. This verification procedure guarantees that the method by Hoffmann et al. only generates true landmarks; however, unsound orderings may be established due to unsound landmark candidates. While the unsound landmarks are pruned after failing the verification test, unsound *orderings* may remain.

Porteous and Cresswell (2002) propose the alternative approximation for *first achievers* of a fact *F* that we use. They consider all first achievers that are *possibly applicable before F* and thus guarantee the correctness of the found landmarks and orderings. They also find disjunctive landmarks. Our method for landmark detection differs from theirs by adding detection of landmarks via domain transition graphs, and detection of additional orderings via restricted relaxed planning graphs. Porteous and Cresswell additionally reason about multiple occurrences of landmarks (if the same landmark has to be achieved, made false again and re-achieved several times during all plans), which we do not.

The approach by Hoffmann et al. (2004) *exploits* landmarks by decomposing the planning task into smaller subtasks, making the landmarks intermediary goals. Instead of searching for the goal of the task, it iteratively aims to achieve a landmark that is minimal with respect to the orderings. In detail, it first builds a landmark graph (with landmarks as vertices and orderings as arcs). Possible cycles are broken by removing some arcs. The sources $S$ of the resulting directed acyclic graph are handed over to a base planner as a disjunctive goal, and a plan is generated to achieve one of the landmarks in $S$. This landmark, along with its incident arcs, is then removed from the landmark graph, and the process repeats from the end state of the generated plan. Once the landmark graph becomes empty, the base planner is asked to generate a plan to the original goal. (Note that even though all goal facts are landmarks and were thus achieved previously, they may have been violated again.)

As a base planner for solving the subtasks any planner can be used; Hoffmann et al. (2004) experimented with FF. They found that the decomposition into subtasks can lead to a more directed search, solving larger instances than plain FF in many domains. However, we found that their method leads to worse average performance on the IPC benchmarks from 1998 to 2006 when using Fast Downward as a base planner (Richter et al., 2008). Furthermore, the method by Hoffmann et al. often produces solutions that are longer than those produced by the base planner, as the disjunctive search control frequently switches between different parts of the task which may have destructive interactions. Sometimes this even leads to dead ends, so that this approach fails on solvable tasks. By contrast, our approach incorporates landmark information while searching for the original goal of the planning task via a heuristic function derived from the landmarks (see next section). As we have recently shown, this avoids the possibility of dead-ends and usually generates better-quality solutions (Richter et al., 2008).

Sebastia et al. (2006) extend the work by Hoffmann et al. by employing a refined preprocessing technique that groups landmarks into consistent sets, minimising the destructive interactions between the sets. Taking these sets as intermediary goals, they avoid the increased plan length experienced by Hoffmann et al. (2004). However, according to the authors this preprocessing is computationally expensive and may take longer than solving the original problem.

Zhu and Givan (2003) propose a technique for finding landmarks by propagating "necessary predecessor" information in a planning graph. Their definition of landmarks encompasses operators that are necessary in any plan (called *action landmarks*), and they furthermore introduce the notion of a *causal landmark* for fact landmarks that are required as a precondition for some operators in every plan. They argue that fact landmarks which are not causal are "accidental" effects and do not warrant being sought explicitly. Their algorithm computes action landmarks and causal fact landmarks at the same time by propagating information during the construction of a relaxed planning graph. An extended variant of their algorithm is also able to infer multiple occurrences of landmarks. Gregory et al. (2004) build on their work to find disjunctive landmarks through symmetry breaking.

Similar to our work, Zhu and Givan (2003) use the causal fact landmarks and action landmarks to estimate the goal distance of a given state. To this end, they treat each fact landmark as a *virtual action* (sets of operators that can achieve the fact landmark) and obtain a distance estimate by bin packing. The items to be packed into bins are the real landmark actions (singletons) and virtual actions, where each bin may only contain elements such that a pairwise intersection of the elements is non-empty. Zhu and Givan employ a greedy algorithm to estimate the minimum number of bins and use this value as distance estimate. Their experimental results are preliminary, however, and do not demonstrate a significant advantage of their method over the FF planner.

## 5. The Landmark Heuristic

The LAMA planning system uses landmarks to calculate heuristic estimates. Since we know that all landmarks must be achieved in order to reach a goal, we can approximate the goal distance of a state $s$ reached by a path (i. e., a sequence of states) $\pi$ as the estimated number of landmarks that still need to be achieved from $s$ onwards. These landmarks are given by

$$L(s, \pi) := (L \setminus Accepted(s, \pi)) \cup ReqAgain(s, \pi)$$

where $L$ is the set of all discovered landmarks, $Accepted(s, \pi)$ is the set of *accepted* landmarks, and $ReqAgain(s, \pi)$ is the set of accepted landmarks that are *required again*, with the following definitions based on a given landmarks graph $(L, O)$ :

$$Accepted(s, \pi) := \begin{cases} \{\psi \in L \mid s \models \psi \text{ and } \nexists(\varphi \to_x \psi) \in O\} & \pi = \langle\rangle \\ Accepted(s_0[\pi'], \pi') \cup \{\psi \in L \mid s \models \psi & \pi = \pi'; \langle o \rangle \\ \text{ and } \forall(\varphi \to_x \psi) \in O : \varphi \in Accepted(s_0[\pi'], \pi')\} \end{cases}$$

$$ReqAgain(s, \pi) := \{\varphi \in Accepted(s, \pi) \mid s \not\models \varphi \\ \text{ and } (s_\star \models \varphi \text{ or } \exists(\varphi \to_{gn} \psi) \in O : \psi \notin Accepted(s, \pi))\}$$

A landmark $\varphi$ is first accepted in a state $s$ if it is true in that state, and all landmarks ordered before $\varphi$ are accepted in the predecessor state from which $s$ was generated. Once a landmark has been accepted, it remains accepted in all successor states. For the initial state, accepted landmarks are all those that are true in the initial state and do not have any predecessors in the landmark graph. An accepted landmark $\varphi$ is *required again* if it is not true in $s$ and (a) it forms part of the goal or (b) it must be true directly before some landmark $\psi$ (i. e., $\varphi \to_{gn} \psi$) where $\psi$ is not accepted in $s$. In the latter case, since we know that $\psi$ must still be achieved and $\varphi$ must be true in the time step before $\psi$, it holds that $\varphi$ must be achieved again. The number $|L(s, \pi)|$ is then the heuristic value assigned to state $s$. Pseudo code for the heuristic is given in Algorithm 3.

The landmark heuristic will assign a non-zero value to any state that is not a goal state, since goals are landmarks that are always counted as *required again* per condition (a) above. However, the heuristic may also assign a non-zero value to a *goal state*. This happens if plans are found that do not obey the reasonable orderings in the landmark graph, in which case a goal state may be reached without all landmarks being accepted.[3] Hence, we need to explicitly test states for the goal condition in order to identify goal states during search.

Note that this heuristic is path-dependent, i. e., it depends on the sequence of states by which $s$ is reached from the initial state. This raises the question of what happens if a state $s$ can be reached via several paths. In LAMA, the heuristic for a state is calculated only once, when it is first reached. An alternative option would be to re-evaluate $s$ each time a new path to $s$ is discovered, taking into account the information of all paths to $s$ known at the time. As Karpas and Domshlak (2009) note, we can calculate the landmarks that are accepted in $s$ given a set of paths $\mathcal{P}$ to $s$ as $Accepted(s, \mathcal{P}) := \bigcap_{\pi \in \mathcal{P}} Accepted(s, \pi)$, since it holds that any landmark that is not achieved along *all* paths $\pi \in \mathcal{P}$ must

---

3. In the special case where $\varphi \to_r \psi$ and $\varphi$ and $\psi$ can become true simultaneously, we could avoid this by accepting both $\varphi$ and $\psi$ at once (Buffet & Hoffmann, 2010), or we could modify our definition of reasonable orderings such that $\varphi \to_r \psi$ does not hold unless $\psi$ must become true strictly after $\varphi$. The general problem that goal states may be assigned a non-zero value, however, still persists even with these modifications.

Global variables:
    $\Pi = \langle \mathcal{V}, s_0, s_\star, O, C \rangle$     ▷ Planning task to solve
    $LG = \langle L, O \rangle$     ▷ Landmark graph of $\Pi$
    *Accepted*     ▷ Landmarks accepted in states evaluated so far

**function** LM_COUNT_HEURISTIC$(s, \pi)$
    **if** $\pi = \langle \rangle$ **then**     ▷ Initial state
        $Accepted(s, \pi) \leftarrow \{ \psi \in L \mid s_0 \models \psi \text{ and } \nexists(\varphi \rightarrow_x \psi) \in O \}$
    **else**
        $\pi' \leftarrow \langle o_1, \ldots, o_{n-1} \rangle$ for $\pi = \langle o_1, \ldots, o_n \rangle$
        $parent \leftarrow s_0[\pi']$     ▷ $Accepted(parent, \pi')$ has been calculated before
        $Reached \leftarrow \{ \psi \in L \mid s \models \psi \text{ and } \forall(\varphi \rightarrow_x \psi) \in O \colon \varphi \in Accepted(parent, \pi') \}$
        $Accepted(s, \pi) \leftarrow Accepted(parent, \pi') \cup Reached$
    $NotAccepted \leftarrow L \setminus Accepted(s, \pi)$
    $ReqGoal \leftarrow \{ \varphi \in Accepted(s, \pi) \mid s \not\models \varphi \text{ and } s_\star \models \varphi \}$
    $ReqPrecon \leftarrow \left\{ \varphi \in Accepted(s, \pi) \mid s \not\models \varphi \text{ and } \exists \psi \colon (\varphi \rightarrow_{gn} \psi) \in O \wedge \psi \notin Accepted(s, \pi) \right\}$
    **return** $|NotAccepted \cup ReqGoal \cup ReqPrecon|$

Algorithm 3: The landmark count heuristic.

be achieved from $s$ onwards. The heuristic value of $s$ can then be derived from this in an analogous way as before.

The landmark heuristic as outlined above estimates the goal distance of states, i.e., the number of operator applications needed to reach the goal state from a given state. To participate in IPC 2008, we made this function cost-sensitive by *weighting* landmarks with an estimate of their minimum cost. Apart from estimating goal distance by counting the number of landmarks that still need to be achieved from a state, we estimate the cost-to-go from a state by the sum of all minimum costs of those landmarks. The cost counted for each landmark is the minimum action cost of any of its first achievers. (Alternative, more sophisticated methods for computing the costs of landmarks are conceivable and are a potential topic of future work.) The heuristic value LAMA assigns to a state is however not its pure cost-to-go estimate, but rather the *sum* of its cost estimate and its distance estimate. By thus accounting for both the costs-to-go and the goal distances of states, this measure aims to balance speed of the search and quality of the plans, and in particular counter-act the problems that may arise from zero-cost operators (see Section 3.3).

We also generate preferred operators along with the landmark heuristic. An operator is preferred in a state if applying it achieves an *acceptable* landmark in the next step, i.e., a landmark whose predecessors have already been accepted. If no acceptable landmark can be achieved within one step, the preferred operators are those which occur in a relaxed plan to a *nearest* acceptable landmark. A nearest landmark in the cost-unaware setting is one that is relaxed reachable with a minimal number of operators, while in the cost-sensitive setting it is a landmark reachable with the cheapest $h_{add}$ cost (see Section 6), where again both cost and distance estimates are taken into account. This nearest landmark can be computed by building a relaxed planning graph or, equivalently, performing a relaxed exploration (which is what LAMA does, see Section 6), and determining the earliest or least costly occurrence of an acceptable landmark in this structure. A relaxed plan to this landmark is then extracted, and the operators in this plan form preferred operators if they are applicable in the current state.

## 6. The Cost-Sensitive FF/add Heuristic

When we first introduced the landmark heuristic (Richter et al., 2008), it proved not to be competitive on its own, compared to established heuristics like the FF heuristic (Hoffmann & Nebel, 2001). However, the *joint use* of the FF heuristic and the landmark heuristic in a multi-heuristic search improved the performance of a planning system, compared with only using the FF heuristic. This is thus the path LAMA follows. The FF heuristic is based on a *relaxation* of the planning task that ignores delete effects, which in FDR tasks translates to allowing state variables to hold several values simultaneously.

The FF heuristic for a state $s$ is computed in two phases: the first phase, or *forward phase*, calculates an estimate for each fact in the planning task of how costly it is to achieve the fact from $s$ in a relaxed task. Concurrently, it selects an operator called *best support* for each fact $F$, which is a greedy approximation of a cheapest achiever (an achiever $a$ of $F$ where the costs of making $a$ applicable and applying it are minimal among all achievers of $F$, when starting in $s$). In the second phase, a *plan* for the relaxed task is constructed based on the best supports for each fact. This is done by chaining backwards from the goals, selecting the best supports of the goals, and then recursively selecting the best supports for the preconditions of already selected operators. The union of these best supports then constitutes the relaxed plan (i. e., for each fact its best support is only added to the relaxed plan once, even if the fact is needed several times as a precondition). The length of the resulting relaxed plan is the heuristic estimate reported for $s$.

The forward phase can be viewed as propagating cost information for operators and facts in a *relaxed planning graph* (Hoffmann & Nebel, 2001). However, this graph does not need to be explicitly constructed to compute the heuristic. Instead, a form of generalised Dijkstra cheapest-path algorithm as described by Liu, Koenig and Furcy (2002) is used in LAMA, which propagates costs from preconditions to applicable operators and from operators to effects. In this method, each operator and fact is represented only once, reducing the time and space requirements from $O(NK)$, where $N$ is the size of the relaxed planning task and $K$ the depth of the relaxed planning graph, to $O(N)$. In order to deal with conditional effects, operators with $n$ effects are split into $n$ operators with one effect each, and the corresponding effect conditions are moved into the preconditions of those operators. If any of those $n$ operators is selected for inclusion in the relaxed plan, the original operator is included instead (again, each operator is included in the relaxed plan only once).

The cost estimate for an operator in the original FF heuristic is its depth in the relaxed planning graph, which in the case of planning with unit-cost operators is equivalent (Fuentetaja, Borrajo, & Linares López, 2009) to propagating costs via the $h_{max}$ criterion (Bonet & Geffner, 2001). The $h_{max}$ criterion estimates the cost of an operator by the *maximum* over the costs of its preconditions, plus the action cost of the operator itself (1 when planning without action costs). The cost of a fact is estimated as the cost of its cheapest achiever, or zero if the fact is true in the current state $s$. While originally proposed for unit-cost planning, this heuristic can be adapted to cost-based planning in a straightforward way by using action costs in the cost propagation phase, and reporting the total cost of the resulting relaxed plan, rather than its length, as the heuristic estimate.

Using other criteria for cost propagation results in variations of the FF heuristic (Bryce & Kambhampati, 2007; Fuentetaja et al., 2009). One variant that has been previously proposed in the litera-ture (Do & Kambhampati, 2003) is to use the $h_{add}$ criterion (Bonet & Geffner, 2001). It is similar to the $h_{max}$ criterion except for estimating the cost of operators via the *sum*, rather than the maximum, of the costs for their preconditions. We will in the following use the term *FF/add* for this variant of the FF heuristic. Independently of us, Keyder and Geffner (2008) implemented the FF/add heuristic which they call $h_a$ in their planner FF($h_a$) at IPC 2008. A formal specification of the FF/add heuristic can be found in their paper. The heuristic function in LAMA is similar to this cost-sensitive FF/add heuristic. However, as with the landmark heuristic, LAMA is not purely guided by action costs, but rather uses both cost and distance estimates equally. This means that during cost propagation, each operator contributes its action cost *plus 1 for its distance*, rather than just its action cost, to the propagated cost estimates.

## 7. Experiments

To evaluate how much each of the central features of LAMA contributes to its performance, we have conducted a number of experiments comparing different configurations of these features. We focus our detailed evaluation on the benchmark tasks from the International Planning Competition (IPC) 2008, as we are interested in the setting of planning with action costs. The effect of landmarks in classical planning tasks without actions costs has been studied in previous work (Richter et al., 2008), but we provide summarising results for this case, using the domains of IPCs 1998–2006, in Section 7.6. The benchmark set of IPC 2008 comprises 9 domains with 30 tasks each, resulting in a total of 270 tasks. For one of the domains (Openstacks), two different formulations were available (STRIPS and ADL). As in the competition, we report the better result of those two formulations for each planner.

As described in Section 1, LAMA builds on the platform provided by Fast Downward in three major ways: (1) through the use of landmarks, (2) by using cost-sensitive heuristics to guide search for cheap plans, and (3) by employing anytime search to continue to search for better solutions while time remains. To examine the usefulness of landmarks, we conduct experiments with and without them, while keeping all other planner features fixed. The use of action costs in LAMA is the result of a number of design decisions. Both the landmark heuristic and the FF/add heuristic have been made cost-sensitive. However, rather than focusing purely on action costs, LAMA uses both distance estimates and cost estimates in combination (see Section 3.3) to balance speed and quality of the search. To measure the benefit of this combining approach, we test three different approaches to dealing with costs: (a) using the traditional cost-unaware heuristics (distance estimates), (b) using purely cost-sensitive heuristics (though using distance estimates for tie-breaking), and (c) using the combination of the distance and cost estimates, as in LAMA. The different choices regarding landmarks and approaches to action costs thus result in the following six planner configurations:

- **F**: Use the cost-unaware FF/add heuristic (estimating goal distance).

- **F$_c$**: Use the purely cost-sensitive FF/add heuristic (estimating cost-to-go).

- **F$_c^+$**: Use the FF/add heuristic that combines action costs and distances.

- **FL**: Use the cost-unaware variants of both the FF/add heuristic and the landmark heuristic.

- **FL$_c$**: Use the purely cost-sensitive variants of both heuristics.

- **FL$_c^+$**: Use the variants that combine action costs and distances for both heuristics.

Note that in contrast to the setting of optimal planning (Karpas & Domshlak, 2009), the landmark heuristic by itself is not competitive in our case, and landmarks in LAMA are used only to provide *additional* information to an already guided search. As such, we do not include any configurations using only landmarks as heuristic estimators in our detailed results. However, we provide summarising results supporting our claim that they are not competitive.

Each configuration is run with iterated (anytime) search. When highlighting the contribution of the iterated search, we report first solutions vs. final solutions, where the final solution of a configuration is the last, and best, solution it finds within the 30-minute timeout. (Note that the quality of a solution is always determined by its cost, irrespective of whether the heuristic used to calculate it is cost-sensitive or not.) When discussing the three possible approaches to costs (cost-unaware search, purely cost-sensitive search, or LAMA's combination of distances and costs) we write $\mathbf{X}$, $\mathbf{X_c}$, and $\mathbf{X_c^+}$ to denote the three cost approaches independently of the heuristics used.

We measure performance using the same criterion that was employed at IPC 2008 (Helmert et al., 2008). Each planner configuration is run for 30 minutes per task. After the timeout, a planner aggregates the ratio $c^*/c$ to its total score if $c$ is the cost of the plan it has found, and $c^*$ is the cost of the best known solution (e. g., a reference solution calculated by the competition organisers, or the best solution found by any of the participating planners).

Experiments were run on the hardware used in the competition, a cluster of machines with Intel Xeon CPUs of 2.66GHz clock speed. The time and memory limits were set to the same values as in the competition, using a timeout of 30 minutes and a memory limit of 2 GB. In the following, we first provide a general overview of the results. Then we discuss special cases, i. e., domains where the results for certain configurations deviate from the overall trend, and try to give plausible explanations for why this may happen.

## 7.1 Overview of Results

In this section, we show that the purely cost-based FF/add configuration $\mathbf{F_c}$ solves significantly fewer tasks than its cost-unaware counterpart $\mathbf{F}$. While $\mathbf{F_c}$ finds *higher-quality* solutions, this does not make up for its low coverage (number of solved tasks) when measuring performance with the IPC criterion. Using *landmarks* improves quality slightly, so that cost-unaware search using landmarks ($\mathbf{FL}$) achieves the highest IPC performance score amongst our configurations. When using the cost-sensitive FF/add heuristic, adding landmarks (resulting in the configurations $\mathbf{FL_c}$ and $\mathbf{FL_c^+}$) increases coverage substantially, while incurring only a small loss in quality. Iterated search improves the scores of all configurations significantly. Lastly, using the combination of cost and distance estimates in the heuristics ($\mathbf{X_c^+}$) is superior to pure cost-based search when using iterated search. Together, using landmarks and the combination of cost and distance estimates ($\mathbf{FL_c^+}$) achieves nearly the same performance as the $\mathbf{FL}$ configuration.

In the following, we support these findings with experimental data. In Section 7.1.1 (*Performance in Terms of the IPC Score*), we show that the cost-sensitive FF/add heuristic by itself scores lowly in terms of the IPC criterion, but that landmarks and the combination of cost and distance estimates together make up for this bad performance. Furthermore, our results demonstrate the magnitude of the impact that iterated search has on the performance scores. In Section 7.1.2 (*Coverage*), we show that the bad performance of the cost-sensitive FF/add heuristic is due to solving fewer tasks, and that the use of landmarks mitigates this problem. In Section 7.1.3 (*Quality*), we present data showing that the purely cost-sensitive FF/add heuristic finds higher-quality plans than the cost-unaware FF/add heuristic in the first search, but that with iterated search, this difference all but disappears. Furthermore, after iterated search the intermediate approach of using cost and distance estimates scores higher than the purely cost-based search. LAMA's approach of using landmarks and the combination of cost and distance estimates ($FL_c^+$) thus effectively mitigates the bad performance of the cost-sensitive FF/add heuristic.

| Domain | Base | **IPC Planners** | | | LAMA | **Slowed LAMA** | | $FL_c^+$ |
|---|---|---|---|---|---|---|---|---|
| | | **C3** | **FF($h_a$)** | **FF($h_a^s$)** | | **×10** | **×100** | |
| Cyber Security | 4 | 9 | 20 | 20 | 28 | 27 | 26 | 28 |
| Elevators | 21 | 16 | 9 | 10 | 20 | 20 | 17 | 22 |
| Openstacks | 21 | 10 | 8 | 8 | 27 | 27 | 26 | 27 |
| PARC Printer | 27 | 18 | 16 | 23 | 21 | 19 | 12 | 22 |
| Peg Solitaire | 20 | 20 | 21 | 23 | 29 | 29 | 26 | 29 |
| Scanalyzer | 24 | 23 | 24 | 24 | 26 | 25 | 22 | 26 |
| Sokoban | 21 | 18 | 15 | 18 | 24 | 22 | 15 | 23 |
| Transport | 18 | 6 | 15 | 14 | 27 | 25 | 21 | 26 |
| Woodworking | 14 | 24 | 22 | 22 | 25 | 24 | 17 | 24 |
| **Total** | **169** | **143** | **150** | **162** | **227** | **218** | **183** | **227** |
| (Total IPC 2008) | (176) | (151) | (157) | (169) | (236) | (—) | (—) | (—) |

| Domain | **First solutions** | | | | | | **Final solutions (iterated search)** | | | | | |
|---|---|---|---|---|---|---|---|---|---|---|---|---|
| | **F** | **$F_c$** | **$F_c^+$** | **FL** | **$FL_c$** | **$FL_c^+$** | **F** | **$F_c$** | **$F_c^+$** | **FL** | **$FL_c$** | **$FL_c^+$** |
| Cyber Security | 20 | 24 | 24 | 20 | 28 | 27 | 23 | 24 | 25 | 26 | 28 | 28 |
| Elevators | 22 | 9 | 9 | 23 | 14 | 16 | 29 | 10 | 15 | 27 | 16 | 22 |
| Openstacks | 20 | 23 | 20 | 13 | 13 | 14 | 27 | 29 | 28 | 27 | 28 | 27 |
| PARC Printer | 20 | 16 | 16 | 23 | 21 | 21 | 23 | 16 | 16 | 24 | 22 | 22 |
| Peg Solitaire | 20 | 23 | 20 | 20 | 22 | 21 | 29 | 29 | 29 | 29 | 29 | 29 |
| Scanalyzer | 19 | 21 | 20 | 22 | 21 | 21 | 24 | 24 | 25 | 29 | 24 | 26 |
| Sokoban | 18 | 20 | 19 | 18 | 19 | 19 | 24 | 24 | 24 | 22 | 23 | 23 |
| Transport | 18 | 15 | 15 | 24 | 24 | 23 | 19 | 17 | 17 | 24 | 26 | 26 |
| Woodworking | 22 | 20 | 20 | 20 | 20 | 20 | 23 | 21 | 22 | 20 | 23 | 24 |
| **Total** | **180** | **171** | **162** | **182** | **183** | **182** | **220** | **194** | **201** | **229** | **217** | **227** |

Table 1: Performance scores (rounded to whole numbers) for planners scoring $\geq 100$ points at IPC 2008 (top) and our 6 experimental configurations (bottom). Scores for IPC planners were re-calculated (see text). LAMA ×10 and ×100 refer to the results achieved by LAMA when slowed down by factors of 10 and 100, respectively. $FL_c^+$ is essentially the same as the IPC planner LAMA.

### 7.1.1 Performance in Terms of the IPC Score

The scores of all planners scoring more than 100 points at IPC 2008 are shown in the top part of Table 1. Apart from LAMA, this includes a base planner run by the competition organisers (FF with a preprocessing step that compiles away action costs), the FF($h_a$) and FF($h_a^s$) planners by Keyder and Geffner (2008) and the C3 planner by Lipovetzky and Geffner (2009). The plans found by these planners have been obtained from the competition website (Helmert et al., 2008). However, the scores for those plans depend on the best known solutions for the tasks. The scores we show here thus differ from the ones published at IPC 2008, as we have re-calculated them to reflect new best solutions found in our experiments. To illustrate the magnitude of the change, the original total scores of the IPC planners are shown in parentheses in the last table row.

While the configuration $\mathbf{FL_c^+}$ results in essentially the same planner as (the IPC version of) LAMA, we report its results again, as some minor corrections have been implemented in LAMA since the competition. In addition, the planner makes arbitrary decisions at some points during execution due to underlying programming library methods, leading to varying results. However, as Table 1 shows these differences between $\mathbf{FL_c^+}$ and LAMA are very small. We have furthermore added columns to the table showing the hypothetical results for LAMA that would be obtained if its search were slowed down by the constant factors 10 and 100, respectively (i. e., the results obtained when cutting of the search after 3 minutes, or 18 seconds, respectively). The numbers show that LAMA still outperforms the other IPC planners even with a severe handicap, demonstrating that the good performance of LAMA is not mainly due to an efficient implementation.

The bottom part of Table 1 contains the results for our six experimental configurations after the first search iteration (left) and after the 30-minute timeout (right). As can be seen, both the use of landmarks and iterated search lead to significant improvements in performance. Even with just one of those two features our planner performs notably better than any of its competitors from IPC 2008. (Note however that the baseline planner performed very badly in Cyber Security due to problems with reading very large task descriptions.) In combination, the benefits of landmarks and iterated search grow further: in cost-unaware search the use of landmarks results in 2 additional score points for the first solutions, but in 9 additional points for the final solutions. Similar results hold for the cost-sensitive configurations. This is mainly due to the Openstacks domain, where using landmarks is highly detrimental to solution quality for the first solutions. Iterated search mitigates the problem by improving quality to similar levels with and without landmarks. Overall, there is thus a slight synergy effect between landmarks and iterated search, making the joint benefit of the two features larger than the sum of their individual contributions. The effect of landmarks in the Openstacks domain is discussed in more detail later.

The use of cost-sensitive search did not pay off in our experiments. Cost-unaware search is always at least roughly equal, and often substantially better than the cost-sensitive configurations. Cost-sensitive planning seems to be not only a problem for LAMA, but also for the other participating planners at IPC 2008: notably, all cost-sensitive competitors of LAMA fare worse than the cost-ignoring baseline. In LAMA, best performance is achieved by using cost-unaware search with landmarks and iterated search. However, using the combination of cost and distance estimates instead ($\mathbf{FL_c^+}$) leads to performance that is almost equally good. In particular, $\mathbf{FL_c^+}$ is substantially better than the pure cost search $\mathbf{FL_c}$ if iterated search is used.

A more detailed view on the same data is provided in Figure 6, where we show the performance over time for our six experimental configurations. A data point at 100 seconds, for example, shows the score the corresponding planner would have achieved if the timeout had been 100 seconds. As the top panel shows, cost-sensitive search is consistently worse than cost-unaware search when using only the FF/add heuristic. Using landmarks (see centre panel), the two settings $\mathbf{FL}$ and $\mathbf{FL_c^+}$ achieve better performance than $\mathbf{F}$, though $\mathbf{FL_c^+}$ needs 2 minutes to surpass $\mathbf{F}$, while $\mathbf{FL}$ does so within 5 seconds. Pure cost search, even with landmarks ($\mathbf{FL_c}$), performs worse than $\mathbf{F}$ at all times. The bottom panel of Figure 6 shows that when not using iterated search, the performance of the 4 best configurations $\mathbf{FL}$, $\mathbf{F}$, $\mathbf{FL_c^+}$, and $\mathbf{FL_c}$ is fairly similar eventually, but the cost-sensitive approaches need more time than the cost-unaware configurations to reach the same performance levels.

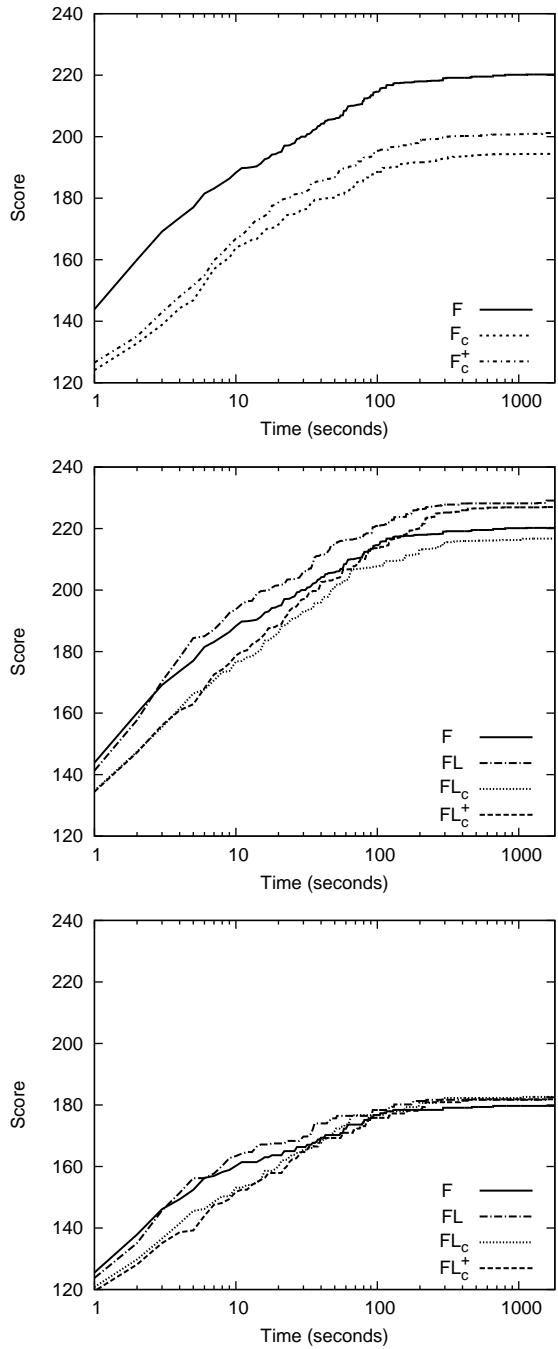

Figure 6: Score over time using iterated search (top and centre panel) and without iterated search, i. e., showing first solutions only (bottom panel).

| Domain | Base | C3 | $\mathbf{FF}(h_a)$ | $\mathbf{FF}(h_a^s)$ | LAMA | $\mathbf{FL_c^+}$ |
|---|---|---|---|---|---|---|
| Cyber Security | 4 | 15 | 23 | 22 | 30 | 30 |
| Elevators | 30 | 30 | 23 | 26 | 24 | 25 |
| Openstacks | 30 | 30 | 25 | 26 | 30 | 30 |
| PARC Printer | 30 | 18 | 16 | 23 | 22 | 23 |
| Peg Solitaire | 30 | 30 | 29 | 29 | 30 | 30 |
| Scanalyzer | 30 | 27 | 28 | 28 | 30 | 30 |
| Sokoban | 27 | 22 | 17 | 20 | 25 | 24 |
| Transport | 29 | 12 | 23 | 22 | 30 | 30 |
| Woodworking | 17 | 28 | 29 | 29 | 30 | 30 |
| **Total** | **227** | **212** | **213** | **225** | **251** | **252** |

| Domain | $\mathbf{F}$ | $\mathbf{F_c}$ | $\mathbf{F_c^+}$ | $\mathbf{FL}$ | $\mathbf{FL_c}$ | $\mathbf{FL_c^+}$ |
|---|---|---|---|---|---|---|
| Cyber Security | 30 | 28 | 29 | 30 | 30 | 30 |
| Elevators | 30 | 15 | 16 | 30 | 22 | 25 |
| Openstacks | 30 | 30 | 30 | 30 | 30 | 30 |
| PARC Printer | 25 | 16 | 16 | 24 | 23 | 23 |
| Peg Solitaire | 30 | 30 | 30 | 30 | 30 | 30 |
| Scanalyzer | 28 | 30 | 29 | 30 | 30 | 30 |
| Sokoban | 25 | 25 | 24 | 23 | 24 | 24 |
| Transport | 26 | 22 | 21 | 29 | 30 | 30 |
| Woodworking | 30 | 28 | 28 | 28 | 30 | 30 |
| **Total** | **254** | **224** | **223** | **254** | **249** | **252** |

Table 2: Coverage for planners scoring $\geq 100$ points at IPC 2008 (top) and our 6 experimental configurations (bottom). Results of the IPC planners have been taken from the competition. $\mathbf{FL_c^+}$ is essentially the same as the IPC planner LAMA.

### 7.1.2 Coverage

The bad performance of cost-sensitive search is surprising, given that our performance criterion awards higher scores to cheaper plans. One explanation could be that this is mainly due to different coverage. If finding plans of high quality is substantially harder than finding plans of low quality, then focusing on nearest goals rather than cheapest goals may solve more tasks within a given time limit. In Table 2 we show the coverage for all considered planners and configurations. The numbers confirm that when not using landmarks, the coverage of cost-unaware search is indeed substantially higher than the coverage of cost-sensitive search. However, with landmarks, the differences in coverage between the various cost approaches are small. In particular, landmarks do not improve coverage further for the cost-unaware search, but bring the cost-sensitive configurations up to the same coverage level as the cost-unaware search. Landmarks thus seem to be very helpful in overcoming the coverage problems of cost-sensitive search.

| Domain | $F_c$ / F | | $F_c^+$ / F | | $FL_c$ / $F_c$ | | $FL_c^+$ / $F_c^+$ | |
|---|---|---|---|---|---|---|---|---|
| | Tasks | C. Ratio | Tasks | C. Ratio | Tasks | C. Ratio | Tasks | C. Ratio |
| Cyber Security | 28 | 0.64 | 29 | 0.69 | 28 | 0.81 | 29 | 0.83 |
| Elevators | 15 | 1.16 | 16 | 1.15 | 14 | 0.89 | 16 | 0.92 |
| Openstacks | 30 | 0.83 | 30 | 1.00 | 30 | 1.98 | 30 | 1.46 |
| PARC Printer | 16 | 0.79 | 16 | 0.79 | 15 | 1.05 | 15 | 1.05 |
| Peg Solitaire | 30 | 0.87 | 30 | 1.02 | 30 | 1.04 | 30 | 0.95 |
| Scanalyzer | 28 | 0.94 | 28 | 0.93 | 30 | 0.98 | 29 | 1.00 |
| Sokoban | 23 | 0.94 | 22 | 0.98 | 24 | 0.98 | 23 | 1.00 |
| Transport | 21 | 1.01 | 21 | 1.00 | 22 | 0.89 | 21 | 0.89 |
| Woodworking | 28 | 1.02 | 28 | 1.02 | 28 | 1.07 | 28 | 1.06 |
| **Total** | 219 | **0.88** | 220 | **0.94** | 221 | **1.06** | 221 | **1.01** |

| Domain | $F_c$ / F | | $F_c^+$ / F | | $FL_c$ / $F_c$ | | $FL_c^+$ / $F_c^+$ | |
|---|---|---|---|---|---|---|---|---|
| | Tasks | C. Ratio | Tasks | C. Ratio | Tasks | C. Ratio | Tasks | C. Ratio |
| Cyber Security | 28 | 0.81 | 29 | 0.82 | 28 | 0.81 | 29 | 0.82 |
| Elevators | 15 | 1.51 | 16 | 1.05 | 14 | 0.89 | 16 | 0.99 |
| Openstacks | 30 | 0.95 | 30 | 0.96 | 30 | 1.04 | 30 | 1.04 |
| PARC Printer | 16 | 0.97 | 16 | 0.97 | 15 | 1.01 | 15 | 1.01 |
| Peg Solitaire | 30 | 0.99 | 30 | 1.00 | 30 | 1.02 | 30 | 1.01 |
| Scanalyzer | 28 | 1.01 | 28 | 0.93 | 30 | 1.02 | 29 | 1.01 |
| Sokoban | 23 | 1.01 | 22 | 1.00 | 24 | 1.02 | 23 | 1.00 |
| Transport | 21 | 0.98 | 21 | 0.89 | 22 | 0.89 | 21 | 0.89 |
| Woodworking | 28 | 0.99 | 28 | 0.94 | 28 | 1.00 | 28 | 1.01 |
| **Total** | 219 | **0.99** | 220 | **0.94** | 221 | **0.97** | 221 | **0.97** |

Table 3: Average ratio of the first solution costs (top) and best solution costs after iterative search (bottom) for various pairs of configurations on their commonly solved tasks.

As mentioned before, the landmark heuristic by itself is however not competitive. Using only the landmark heuristic and not the FF/add heuristic results in IPC 2008 performance scores between 164 and 167 with iterated search, and coverage points between 185 and 189 for the three possible cost settings. This is substantially worse then the performance scores greater than 194 and coverage points greater than 223 achieved by any of the other LAMA configurations.

### 7.1.3 Quality

As a next step, we look purely at solution quality. Firstly, we want to answer the question whether the improvement in coverage achieved by landmarks in the cost-sensitive search comes at a price in solution quality, i. e., whether using landmarks directs the search to close goals rather than cheap goals. Secondly, we would like to know how the solution quality differs between the cost-sensitive and the cost-unaware configurations. In particular, how much quality do we lose by combining distance and cost estimates ($\mathbf{X_c^+}$) as opposed to using pure cost search ($\mathbf{X_c}$)? The score used at IPC 2008 and in Table 1 incorporates both coverage and quality information by counting unsolved tasks as 0 – a method that allows ranking several planners solving different subsets of the total benchmark set. When we are interested in examining quality independent of coverage, we must restrict our focus on those tasks solved by all compared planners. Table 3 contains quality information comparing the solution costs of several configurations, where we compare configurations pair-wise in order to maximise the number of commonly solved tasks. The top part of Table 3 contains comparisons involving the *first* solutions found by each configuration, while the bottom part of the table concerns the *best* solutions found after iterative search. For each pair of configurations we show the number of tasks solved by both, and the geometric mean of the *cost ratio* for the plans they find.

As expected, the cost-sensitive configurations $\mathbf{F_c}$ and $\mathbf{F_c^+}$ find cheaper plans than the cost-unaware configuration $\mathbf{F}$ on average, where in particular the pure cost search $\mathbf{F_c}$ finds high-quality *first* plans (see the first column in the top part of the table). For both $\mathbf{F_c}$ and $\mathbf{F_c^+}$, however, the difference to $\mathbf{F}$ is not very large. In some domains, most notably in Elevators, the plans found by the cost-sensitive heuristics are actually *worse* than the plans found by cost-unaware search.

Landmarks deteriorate quality for the first plans of $\mathbf{F_c}$; but $\mathbf{F_c^+}$, which starts out with a worse quality than $\mathbf{F_c}$, is not noticeably further deteriorated by landmarks. For both configurations, however, the main negative impact through landmarks is in the Openstacks domain, where plans become nearly twice as expensive for $\mathbf{F_c}$, and 50% more expensive for $\mathbf{F_c^+}$. By contrast, in the remaining 8 domains average plan quality for both configurations with landmarks is even slightly *better* on average than without landmarks.

We note that iterative search has a remarkable impact on the relative performance of the different configurations. When looking at the solutions found after iterative search, $\mathbf{F_c}$ actually performs worse than $\mathbf{F_c^+}$, whereas it is the other way round for the first solutions (compare the first two columns in the top row versus the bottom row of the table). This can be explained to some extent by the fact that the same reasons that cause $\mathbf{F_c}$ to have low coverage also prevent it from improving much over time. As we will show in selected domains later, the cost-sensitive heuristic often expands many more nodes than the cost-unaware search, leading to the observed behaviour. This is most likely due to the fact that finding plans of high quality is hard and thus unsuccessful in many of the benchmark tasks. For example, in some domains cost-sensitive search leads to large local minima that do not exist for cost-unaware search. More generally, good plans are often longer than bad plans, which may lead to increased complexity in particular in domains where the heuristic values are inaccurate. We will showcase the problems of cost-sensitive search in more detail in the Elevators and PARC Printer domains later on.

With iterative search, landmarks do not deteriorate quality for either $\mathbf{F_c}$ nor $\mathbf{F_c^+}$ on average, as the negative impact of the Openstacks domain is no longer present. (This effect in the Openstacks domain will be discussed in more detail later.)

Summarising our findings, we can say that landmarks effectively support the cost-sensitive FF/add heuristic in finding solutions, without steering the search away from good solutions. Similarly, combining distance and cost estimates as in $\mathbf{X_c^+}$ leads the search to finding solutions quickly without overly sacrificing quality, as is demonstrated by its superior anytime performance compared to pure cost search.

By way of example, we now present detailed results for four of the nine competition domains. We choose domains that we deem to be of particular interest because the results in them either exaggerate or contradict the general trends discussed so far. The domains Elevators and PARC Printer highlight the problems of cost-sensitive search; in Cyber Security cost-sensitive search performs uncharacteristically well; and Openstacks is a domain where landmarks do not lead to the usual improvement, but rather to a deterioration of performance.

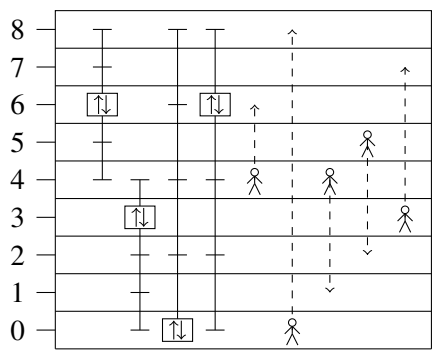

Figure 7: An example elevators task.

## 7.2 Elevators

The Elevators domain models the transportation of passengers in a building via fast and slow elevators, where each elevator has a certain passenger capacity and can access certain floors. Passengers may have to change elevators to get to their final destination, and furthermore the two different types of elevators have different associated cost functions. This is in contrast to the Miconic domain, used in an earlier international planning competition (Bacchus, 2001), which also models the transporting of passengers via elevators, but where there is only one elevator that can access all floors with just one (unit-cost) operator. In Elevators, the floors in the building are grouped into blocks, overlapping by one floor. Slow elevators only operate within a block and can access all floors within their block. Fast elevators can access all blocks, but only certain floors within each block (in the first 10 IPC tasks every second floor, and in the other 20 tasks every fourth floor). Fast elevators are usually more expensive than slow elevators except for a distance of two floors, where both elevator types cost the same. However, fast elevators may sometimes be advantageous when transporting passengers between blocks (as they avoid the need for passengers to switch elevators on the shared floor between blocks), and they usually have a higher capacity.

An example task with eight floors, grouped into two blocks, is shown in Figure 7. There are a total of four elevators, two slow ones and two fast ones. The cost function used in the 30 IPC tasks for moving an elevator from its current location to a target floor is $6 + n$ for slow elevators and $2 + 3n$ for fast elevators, where $n$ is the distance travelled (the number of floors between the current location of the elevator and its target). Operators concerning passengers boarding and leaving elevators are free of cost. Assuming this cost function, it is cheaper in this example to transport the passenger located at floor 0 using the two slow elevators (changing at floor 4) than using a direct fast elevator.

Elevators is one of the domains where configurations using the cost-sensitive FF/add heuristic solve far fewer problems than their cost-unaware counterparts. Using landmarks increases coverage, but does not solve the problem completely. Furthermore, it is notable that on the problems that the cost-sensitive configurations do solve, their solutions often have *worse quality* than the solutions of the cost-unaware configurations. Table 4 illustrates this fact for the first solutions found when using only the FF/add heuristic. With iterative search (not shown), the solution quality for $F_c^+$ improves to a similar level as that of $F$, whereas $F_c$ remains substantially worse.

| | Quality (IPC Score) | | | Length | | |
|------|------|------|------|------|------|------|
| Task | F | $F_c$ | $F_c^+$ | F | $F_c$ | $F_c^+$ |
| 01 | 0.57 | **0.59** | 0.53 | 26 | 24 | 27 |
| 02 | 0.69 | **0.72** | **0.72** | 27 | 25 | 25 |
| 03 | **0.88** | 0.58 | 0.51 | 21 | 41 | 42 |
| 04 | 0.71 | 0.70 | **0.72** | 34 | 45 | 47 |
| 05 | **0.68** | 0.54 | 0.54 | 33 | 50 | 50 |
| 06 | 0.60 | 0.60 | **0.61** | 56 | 64 | 53 |
| 07 | 0.38 | **0.46** | 0.40 | 71 | 81 | 83 |
| 08 | **0.84** | 0.54 | 0.51 | 47 | 62 | 65 |
| 09 | **0.71** | 0.54 | 0.57 | 54 | 81 | 59 |
| 11 | **0.66** | 0.52 | 0.52 | 39 | 51 | 47 |
| 12 | **0.70** | 0.54 | 0.58 | 55 | 79 | 79 |
| 13 | **0.58** | 0.51 | 0.54 | 60 | 84 | 72 |
| 14 | 0.70 | 0.70 | 0.70 | 81 | 101 | 95 |
| 20 | **0.67** | 0.47 | 0.58 | 132 | 173 | 154 |
| 21 | 0.70 | 0.63 | **0.71** | 84 | 83 | 82 |
| **Avg.** | **0.67** | 0.58 | 0.58 | 55 | 67 | 65 |

Table 4: Comparison of plan qualities (measured via the IPC scores) and plan lengths for the first solutions of $F$, $F_c$, and $F_c^+$ in Elevators. Shown are all tasks solved by all three configurations, with bold print indicating the best solution.

While we do not have a full explanation for why the configurations involving the cost-sensitive FF/add heuristic perform so badly in this domain, several factors seem to play a role. Firstly, in its attempt to optimise costs, the cost-sensitive FF/add heuristic focuses on relatively complex solutions involving mainly slow elevators and many transfers of passengers between elevators, where the relaxed plans are less accurate (i. e., translate less well to actual plans), than in the case for the cost-unaware heuristic. Secondly, the costs associated with the movements of elevators dominate the heuristic values, causing local minima for the cost-sensitive heuristic. Thirdly, the capacity constraints associated with elevators may lead to plateaus and bad-quality plans in particular for the cost-sensitive heuristic. In the following sections, we describe each of these factors in some detail.

Lastly, we found that the *deferred heuristic evaluation* technique used in LAMA (see Section 3.3) did not perform well in this domain. When not using deferred evaluation, the $F_c$ configuration solves 3 additional tasks (though the quality of solutions remains worse than with the $F$ configuration). This partly explains why the FF($h_a$) planner by Keyder and Geffner (2008) has a substantially higher coverage than our $F_c$ configuration in this domain. While the two planners use the same heuristic, they differ in several aspects. Apart from deferred evaluation these aspects include the search algorithm used (greedy best-first search vs. enhanced hill-climbing) and the method for using preferred operators (maintaining additional queues for preferred states vs. pruning all non-preferred successor states).

|       | Slow moves | Fast moves | Ratio fast/slow |
|-------|------------|------------|-----------------|
| $\mathbf{F}$ | 275 | 45 | 6.11 |
| $\mathbf{F_c}$ | 405 | 21 | 19.29 |
| $\mathbf{F_c^+}$ | 404 | 12 | 33.67 |

Table 5: Total elevator moves and ratio of fast/slow moves in the first solutions found by the $\mathbf{F}$, $\mathbf{F_c}$, and $\mathbf{F_c^+}$ configurations, on the 15 Elevators instances solved by all three configurations.

### 7.2.1 Slow vs. Fast Elevators

When examining the results, we found that the $\mathbf{F_c}$ and $\mathbf{F_c^+}$ configurations tend to produce plans where slow elevators are used for most or all of the passengers, while the $\mathbf{F}$ configuration uses fast elevators more often (cf. Table 5). This is not surprising, as for each individual passenger, travelling from their starting point to their destination tends to be *cheaper* in a slow elevator (unless the distance is very short), whereas *fewer operators* are typically required when travelling in a fast elevator. The independence assumptions inherent in the FF/add heuristic (see Section 6) lead to constructing relaxed plans that aim to optimise the transportation of each passenger individually, rather than taking synergy effects into account.

The plans produced by $\mathbf{F_c}$ and $\mathbf{F_c^+}$ are also longer, on average, than the plans produced by $\mathbf{F}$ (see Table 4), one reason for this being that the predominant use of slow elevators requires passengers to change between elevators more often. As plans become longer and involve more passengers travelling in each of the slow elevators, heuristic estimates may become worse. For example, the relaxed plans extracted for computation of the heuristic are likely to abstract away more details if more passengers travel in the same elevator (e. g., since once a passenger has been picked up from or delivered to a certain location, the elevator may "teleport" back to this location with no extra cost in a relaxed plan to pick up or deliver subsequent passengers). Generally, we found that the relaxed plans for the *initial state* produced by $\mathbf{F_c}$ and $\mathbf{F_c^+}$ tend to be similar in length and cost to those produced by $\mathbf{F}$, but the final *solutions* produced by $\mathbf{F_c}$ and $\mathbf{F_c^+}$ are worse than those of $\mathbf{F}$. One reason for this is probably that the increased complexity of planning for more passenger change-overs between elevators in combination with worse relaxed plans poses a problem to the cost-sensitive FF/add heuristic.

### 7.2.2 Local Minima Due to Elevator-Movement Costs

Since action costs model distances, the total cost of a relaxed plan depends on the target floors relative to the current position of an elevator. For both $\mathbf{F_c}$ and $\mathbf{F_c^+}$, the action costs of moving the elevator usually dominate the estimates of the FF/add heuristic. Consider the two example tasks in Figure 8, which differ only in the initial state of the elevators. The elevators need to travel to all three floors in a solution plan, but due to abstracted delete effects a relaxed plan for the initial state will only include operators that travel to the two floors other than the starting floor of the elevator (i. e., the elevator can be "teleported" back to its starting floor without cost). In the left task, the relaxed cost of visiting all three floors is lower than in the right task, as the cost is in the left task is the sum of going from floor 4 to floor 8, and going from floor 4 to floor 0, resulting in a total cost of $10 + 10 = 20$. In the right task, the relaxed cost for visiting all floors is the cost of going from floor 0 to floor 4, and from floor 0 to floor 8, resulting in a total cost of $10 + 14 = 24$. In the left task, once the passenger has boarded the elevator on floor 4, *all immediate successor states have a*

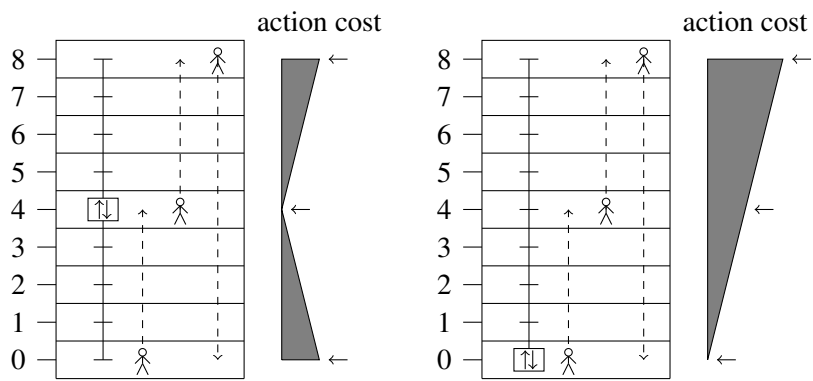

Figure 8: Action cost effects in Elevators in a relaxed setting. Travelling 4 floors costs 10, while travelling 8 floors costs 14. Both tasks have the same solution cost (34), but the left task has a lower relaxed cost (20) than the right task (24).

*worse heuristic estimate due to the movement costs of the elevator.* In particular, the correct action of moving the elevator up to floor 8 (to deliver the passenger) results in a state of worse heuristic value. If we increased the number of waiting passengers at floor 4, the planning system would therefore try boarding all possible subsets of passengers before moving the elevator. And even once the elevator is moved up to floor 8, the heuristic estimate will only improve after the passenger has been dropped off *and* either (a) the elevator has moved back to floor 4, or (b) the second passenger has boarded and the elevator has moved down to floor 0.

Consequently, movement costs may dominate any progress obtained by transporting passengers for a number of successive states. In other words, the planner often has to "blindly" achieve some progress *and* move the elevators towards a middle position given the remaining target floors, in order for the cost-sensitive heuristic to report progress. For the cost-unaware heuristic, the situation is less severe, as the number of elevator movements in the relaxed plan does not increase, and hence the planner encounters a plateau in the search space rather than a local minimum. The use of preferred operators may help to escape from the plateau relatively quickly, whereas a local minimum is much harder to escape from. Two approaches exist that may circumvent this problem. Firstly, the use of enforced hill-climbing (Hoffmann & Nebel, 2001) rather than greedy best-first search is likely to avoid exploration of the entire local minima: in this approach, a breadth-first search is conducted from the first state of a minima/plateau until an improving state is found. Secondly, an improved heuristic could be used that approximates the optimal relaxed cost $h^+$ more exactly. The cost minima shown in Figure 8 is brought about by the independence assumptions inherent in the FF/add heuristic, which estimate the relaxed cost for each goal fact individually in the cheapest possible way. An *optimal* relaxed plan, however, costs the same in the left task as in the right task. A more accurate approximation of the optimal relaxed cost $h^+$ could therefore mitigate the described cost minima. Keyder and Geffner (2009) have recently proposed such an improvement of the FF/add heuristic[4] and shown it to be particularly useful in the Elevators and PARC Printer domains.

---

4. In Keyder & Geffner's approach, the relaxed plan extracted by the FF/add heuristic is improved by iteratively (1) selecting a fact $F$, (2) fixing all operators that are not related to $F$ (because they do not contribute to achieving $F$ nor rely on its achievement), and (3) computing a cheaper way of achieving $F$ given the operators that were fixed in the previous step.

### 7.2.3 PLATEAUS DUE TO CAPACITY CONSTRAINTS

In general, the relaxed plans in the Elevators domain are often of bad quality. One of the reasons is the way the *capacity* of elevators is encoded in the operators for passengers boarding and leaving elevators. For any passenger $p$ transported in an elevator $e$, one of the preconditions for $p$ leaving $e$ is that $n$ passengers be boarded in $e$, where $n$ is a number greater than 0. When constructing a relaxed plan, the FF/add heuristic recursively selects operators that achieve each necessary precondition in the cheapest way. This results in boarding *the passenger that is closest to $e$ in the initial state*, even if this passenger $p'$ is different from $p$, to achieve the condition that some passenger is boarded. The relaxed plan will then contain operators for both boarding $p$ and $p'$ into $e$, and may furthermore contain other operators for boarding $p'$ into whatever elevator $e'$ is deemed best for transporting $p'$. Hence, the relaxed plans often contain many unnecessary boarding operators.

As mentioned in Section 3.3, the greedy best-first search in LAMA breaks ties between equally promising operators by trying the cheaper operator first. Consequently, the zero-cost operators for passengers boarding and leaving elevators are tried first in any state. We found that as soon as one passenger is boarded into a certain elevator, the relaxed plans in the next state are often substantially different, in that more passengers are being assigned to that same elevator. This can be explained by the fact that as soon as one passenger is in an elevator, the precondition for leaving that elevator which is having at least one person boarded, is fulfilled (rather than incurring additional cost). In some example tasks we examined, we found that this effect results in committing to bad boarding operators: LAMA may initially try some bad boarding operator, e. g. boarding the nearest passenger into an elevator to satisfy a capacity precondition for another passenger, as described above. The relaxed plan in the successor state then assigns more passengers to this elevator, at lower cost. Due to the improved heuristic value of the successor state, LAMA retains this plan prefix, even though the first operator was a bad one. It is plausible (though we did not explore it experimentally) that this effect is stronger for the configurations involving the cost-sensitive heuristic, as the costs of their relaxed plans vary more strongly from one state to the next.

More importantly, the capacity constraints lead to plateaus in the search space, as correct boarding and leaving operators are often not recognised as good operators. For example, if the capacity of an elevator is $c$, then boarding the first $c-1$ passengers that need to be transported with this elevator usually leads to improved heuristic values. However, boarding the $c$-th passenger does not result in a state of better heuristic value if there are any further passengers that need to be transported via the same elevator, because the $c$-th passenger boarding destroys the precondition that there must be room in the elevator for other passengers to board. Similarly, the correct leaving of a passenger may not lead to an improved heuristic value if it makes the elevator empty and other passengers need to be transported with that elevator later (because the last passenger leaving destroys the precondition for leaving that there must be at least one passenger boarded).

These effects exist for both the cost-sensitive and the cost-unaware heuristic. However, they typically occur within the plateaus ($\mathbf{F}$) or local minima ($\mathbf{F_c}$, $\mathbf{F_c^+}$) created by the *elevator positions*, as described in the previous section, which means they affect the cost-sensitive configurations more severely. The plateaus become particularly large when several passengers are waiting on the same floor, e. g. when passengers are accumulating on the floor shared by two blocks in order to switch elevators. The planner then tries to board all possible subsets of people into all available elevators (as the zero-cost boarding and leaving operators are always tried first), moving the elevators and even dropping off passengers at other floors, and may still fail to find a state of better heuristic value.

|  |  | Original tasks | | | No capacity constraints | | |
|---|---|---|---|---|---|---|---|
|  |  | **Solved** | **qual. > F** | **qual. < F** | **Solved** | **qual. > F** | **qual. < F** |
| **First plans** | $\mathbf{F_c}$ | 15 | 3 | **10** | 29 | **17** | 6 |
|  | $\mathbf{F_c^+}$ | 16 | 5 | **10** | 30 | **22** | 3 |
| **Final plans** | $\mathbf{F_c}$ | 15 | 0 | **15** | 29 | 9 | **20** |
|  | $\mathbf{F_c^+}$ | 16 | 5 | **6** | 30 | **15** | 11 |

Table 6: Relative qualities of solutions in the original Elevators domain and in a modified variant of the domain where elevators have unlimited capacity. Shown is the total number of tasks solved by the cost-sensitive configurations $\mathbf{F_c}$ and $\mathbf{F_c^+}$, as well as the number of tasks where these configurations find a better/worse plan than the cost-unaware configuration $\mathbf{F}$.

When examining the number of states in local minima for each of the configurations, we found that $\mathbf{F_c}$ and $\mathbf{F_c^+}$ indeed encounter many more such states than $\mathbf{F}$. For example, the percentage of cases in which a state is worse than the best known state is typically around 10% (in rare cases 25%) for $\mathbf{F}$. For $\mathbf{F_c}$ and $\mathbf{F_c^+}$, on the other hand, the numbers are usually more than 35%, often more than 50%, and in large problems even up to 80%.

To verify that the capacity constraints indeed contribute to the bad performance of the cost-sensitive heuristic in this domain, we removed these constraints from the IPC tasks and ran the resulting problems with the $\mathbf{F}$, $\mathbf{F_c}$ and $\mathbf{F_c^+}$ configurations. Not surprisingly, the tasks become much easier to solve, as elevators can now transport all passengers at once. More interestingly though, the bad plan qualities produced by the cost-sensitive configurations (relative to the cost-unaware configuration) indeed become much less frequent, as Table 6 shows.

In summary, our findings suggest that the bad performance of the cost-sensitive FF/add heuristic in the Elevators domain is due to bad-quality relaxed plans (brought about by the focus on slow elevators and the capacity constraints) and plateaus and local minima in the search space (resulting from the movement costs of elevators and the capacity constraints).

### 7.3 PARC Printer

The PARC Printer domain (Do, Ruml, & Zhou, 2008) models the operation of a multi-engine printer capable of processing several printing jobs at a time. Each sheet that must be printed needs to pass through several printer components starting in a feeder and then travelling through transporters, printing engines and possibly inverters before ending up in a finishing tray. The various sheets belonging to a print job must arrive in the correct order at the same finisher tray, but may travel along different paths using various printing engines. There are colour printing engines and ones that print in black and white, where colour printing is more expensive. The action costs of operators are comparatively large, ranging from 2000 to more than 200,000. Colour-printing is the most expensive operator, while operators for printing in black and white cost roughly half as much, and operators for transporting sheets are relatively cheap.

Like in the Elevators domain, the cost-sensitive FF/add heuristic did not perform well here, with $\mathbf{F_c}$ and $\mathbf{F_c^+}$ failing to solve many of the tasks that the cost-unaware configuration $\mathbf{F}$ is able to solve. (Note that $\mathbf{F_c}$ and $\mathbf{F_c^+}$ perform very similarly in this domain, as the large action costs outweigh the distance estimates in $\mathbf{F_c^+}$.) However, in contrast to the Elevators domain, the $\mathbf{F_c}$ and $\mathbf{F_c^+}$ configurations result in notably improved plan quality compared to $\mathbf{F}$. An overview of the number of problems solved and the average quality of first solutions is shown in Table 7. When using landmarks, the differences between cost-sensitive and cost-unaware configurations are strongly reduced, with all three landmark configurations achieving a better performance than the **F** configuration.

|  | **F** | **F$_c^+$** | **FL** | **FL$_c^+$** |
|---|---|---|---|---|
| Tasks solved out of 30 | 25 | 16 | 24 | 23 |
| Avg. quality of first solution | 0.79 | 1.00 | 0.93 | 0.95 |
| Avg. quality of final solution | 0.96 | 1.00 | 1.00 | 0.99 |

Table 7: Coverage vs. quality in the PARC Printer domain. Average qualities are average IPC scores calculated only on those tasks solved by all configurations.

Like in Elevators, we found the quality of relaxed plans to be poor. In the cost-unaware case, a relaxed plan transports sheets from a feeder to the finishing tray via a shortest path, irrespective of whether a suitable printing engine lies on this path. As any path from feeder to finishing tray passes through some printing engine, this frequently involves printing a wrong image on a paper, while additional operators in the relaxed plan handle the transportation from a feeder to a suitable printing engine to print the correct image on the sheet as well. When the cost-sensitive heuristic is used, relaxed plans furthermore become substantially longer, using many transportation operators to reach a cheap printing engine. Analogously to the Elevators domain, the increased complexity associated with longer plans (in combination with the bad quality of the relaxed plans) is thus likely to be the reason for the bad performance of the cost-sensitive heuristic. However, landmarks mitigate the problem, as the numbers of solved tasks in Table 7 clearly show. Landmarks found in this domain encompass those for printing a correct image on each sheet, where a disjunctive landmark denotes the possible printers for each sheet. This helps to counteract the tendencies of the cost-sensitive FF/add heuristic to transport sheets to the wrong printers.

In summary, PARC Printer is like Elevators a domain where the cost-sensitive FF/add heuristic performs badly, though in contrast to Elevators the problem here is purely one of coverage, not of solution quality. Even more than in Elevators, landmarks overcome the problems of the cost-sensitive configurations, improving them to a similar performance levels as the cost-unaware configurations.

### 7.4 Cyber Security

The Cyber Security domain stands out as a domain where the cost-sensitive configurations perform significantly *better* than their cost-unaware counterparts, especially when looking at first solutions. (Iterative search reduces the gap, but does not close it completely.) The domain models the vulnerabilities of computer networks to insider attacks (Boddy, Gohde, Haigh, & Harp, 2005). The task consists in gaining access to sensitive information by using various malware programs or physically accessing computers in offices. Action costs model the likelihood of the attack to fail, i.e., the risk of being exposed. For example, many actions in the office of the attacker, like using the computer, do not involve any cost, whereas entering other offices is moderately costly, and directly instructing people to install specific software has a very high associated cost. In particular, action costs are used to model the desire of finding different methods of attack for the same setting. For example, several tasks in the domain differ only in the costs they associate with certain operators.

In the Cyber Security domain, taking action costs into account pays off notably: while the **F$_c$** and **F$_c^+$** configurations solve 2 and 1 problems less, respectively, than the **F** configuration (see Table 2), they nevertheless result in a better total score. Using landmarks, both cost-sensitive configurations are improved such that they solve all problems while maintaining the high quality of solutions, resulting in an even larger performance gap between $FL_c$ (27.59 points) and $FL_c^+$ (26.60 points) on the one side, and $FL$ (20.43 points) on the other side.

|                                | F     | FL    | $F_c^+$ | $FL_c^+$ |
|--------------------------------|-------|-------|---------|----------|
| IPC score for first solutions  | 20.44 | 20.43 | 23.67   | 26.60    |
| IPC score for final solutions  | 23.12 | 25.93 | 24.69   | 27.53    |

Table 8: IPC scores in the Cyber security domain.

The plans found by the cost-unaware search often involve physically accessing computers in other offices or sending viruses by email, and as such result in large cost. Lower costs can be achieved by more complex plans making sophisticated use of software. As opposed to the Elevators and PARC Printer domains, the relaxed plans in Cyber Security are of very good quality. This explains why the performance of the cost-sensitive heuristic is not negatively impacted by longer plans. Using iterative search improves the performance of $FL$ and $F$ to nearly the same levels as their cost-sensitive counterparts (see Table 8).

### 7.5 Openstacks

The Openstacks domain models the combinatorial optimisation problem *minimum maximum simultaneous open stacks* (Fink & Voß, 1999; Gerevini, Haslum, Long, Saetti, & Dimopoulos, 2009), where the task is to minimise the storage space needed in a manufacturing facility. The manufacturer receives a number of orders, each comprising a number of products. Only one product can be made at a time, and the manufacturer will always produce the total required quantity of a product (over all orders) before beginning the production of a different product. From the time the first product in an order has been produced to the time when all products in the order have been produced, the order is said to be *open* and requires a *stack* (a temporary storage space). The problem consists in ordering the products such that the maximum number of stacks open at any time is minimised. While it is easy to find *a* solution for this problem (any product order is a solution, requiring *n* stacks in the worst case where *n* is the number of orders), finding an *optimal* solution is NP-hard. The minimisation aspect is modelled in the planning tasks via action costs, in that only the operator for opening new stacks has a cost of 1, while all other operators have zero cost. This domain was previously used at IPC 2006 (Gerevini et al., 2009). While that earlier formulation of the domain has unit costs, it is equivalent to the cost formulation described above. Since the number of operators that do not open stacks is the same in every plan for a given task, minimising plan length is equivalent to minimising action costs.

We noticed that in this domain using landmarks resulted in plans of substantially worse quality, compared to not using landmarks. In particular, this is true for the first plans found, whereas the use of anytime search improves the results for both configurations to similar levels. Across all cost settings, using the landmark heuristic in combination with the FF/add heuristic typically produces plans where the majority of orders is started very early, resulting in a large number of simultaneously open stacks, whereas using only the FF/add heuristic leads to plans in which the products corresponding to open orders are manufactured earlier, and the starting of new orders is delayed until earlier orders have been shipped. This is mainly due to the fact that no landmarks are found by LAMA regarding the opening of stacks, which means that due to the choice of action costs in this domain, all landmarks have cost zero and the landmark heuristic is not able to distinguish between plans of different cost. The landmarks found by LAMA relate to the starting and shipping of orders as well as the making of products.[5] However, even if landmarks regarding the opening of stacks were found, they would not be helpful: landmarks state that certain things must be achieved, not that certain things need not be achieved. Landmarks can thus not be used to limit the number of open stacks. The landmark orderings are furthermore not helpful for deciding an order between products, as all product orders are possible—which means that no natural orderings exist between the corresponding landmarks—and no product order results in the form of "wasted effort" captured by reasonable landmark orderings.

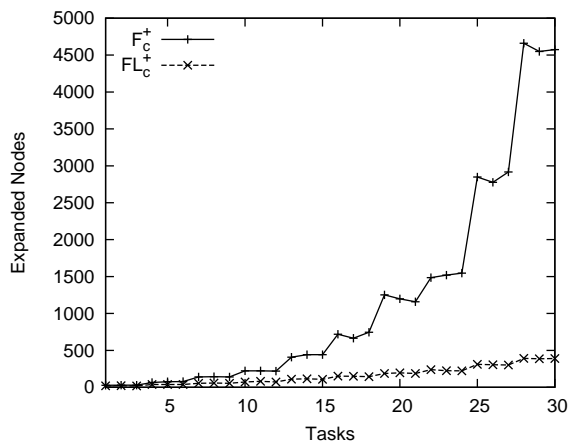

Figure 9: Number of expanded search nodes with and without landmarks in the first search iteration (best-first search) in the Openstacks domain.

As mentioned above, all landmarks found by LAMA have a minimal cost of zero. Therefore, the landmark heuristic fails to estimate the cost to the goal, and distinguishes states only via the *number* of missing started or shipped orders and products. (These goal distance estimates are used directly in **FL**, combined with the all-zero landmark heuristic cost estimates in $\mathbf{FL_c^+}$, and as tie-breakers amongst the zero-cost estimates in $\mathbf{FL_c}$, resulting in the same relative ranking of states by the landmark heuristic in all three cases.) As soon as one stack is open, for each order $o$ the operator that starts $o$ achieves a landmark that is minimal with respect to landmark orderings (namely the landmark stating that $o$ must be started), and the planner thus tends to start orders as soon as possible. The landmark heuristic is not able to take into account future costs that arise through bad product orderings. This is also a problem for the FF/add heuristic, albeit a less severe one: the FF/add heuristic accounts for the cost of opening (exactly) one new stack whenever at least one more stack is needed, and the heuristic will thus prefer states that do not require any further stacks.

The landmark heuristic does, however, provide a good estimate of the goal distance. Since the landmark heuristic prefers states closer to a goal state with no regard for costs, its use results in plans where stacks are opened as needed. This is reflected in our empirical results, where the additional use of the landmark heuristic drastically reduces the number of expanded search nodes (see Figure 9), but leads to higher-cost plans (see Figure 10). Without iterative search, the LAMA configuration $\mathbf{FL_c^+}$ only achieves 13.85 points for this domain, compared to 19.77 points when not using landmarks (configuration $\mathbf{F_c^+}$).

---

5. If the size of disjunctions were not limited in LAMA, it would always find a landmark *stacks_avail*(1) ∨ *stacks_avail*(2) ∨ · · · ∨ *stacks_avail*(n) stating that at least one of the $n$ stacks must be open at some point. However, any landmark stating that two or more stacks need to be open would require a more complex form of landmarks involving conjunction, which LAMA cannot handle.

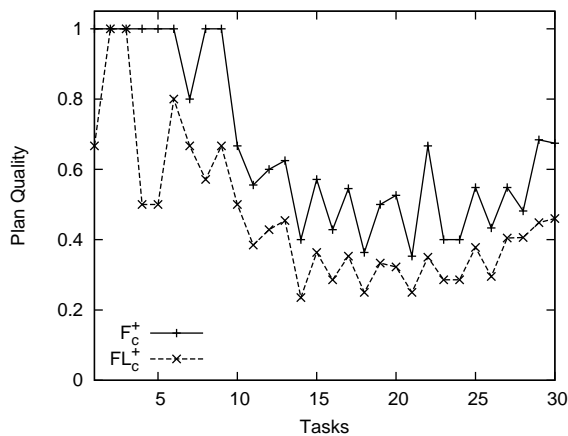

Figure 10: Plan quality (measured via the IPC scores) with and without landmarks in the first search iteration (best-first search) in the Openstacks domain.

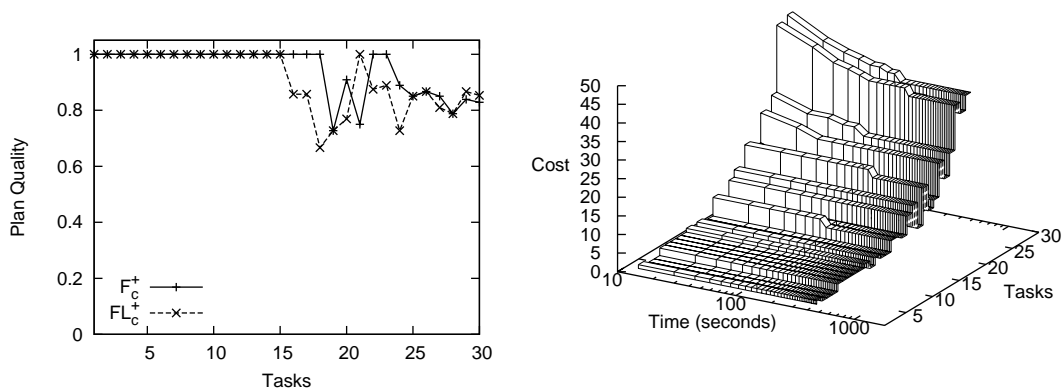

Figure 11: Effect of iterative search in the Openstacks domain. Left: plan quality (IPC score) of the best plan found within 30 minutes with and without landmarks. Right: evolution of plan costs with landmarks ($\mathbf{FL_c^+}$) over time.

Using iterative search, the negative effect of the landmarks on quality is mitigated, as can be seen in Figure 11. $\mathbf{FL_c^+}$ generates up to 21 distinct, and each time improved, plans per problem. In the end, the difference in points is merely 27.40 for $\mathbf{FL_c^+}$ vs. 28.30 for $\mathbf{F_c^+}$. This score is reached after less than 5 minutes of iterated search per task.

Thus, Openstacks is an example for a domain where landmarks are detrimental to solution quality. However, using landmarks provides the benefit of speeding up planning by reducing the number of expanded nodes. This allows iterative search to effectively improve solution quality in the given time limit such that the final results of using landmarks are similar to those of not using landmarks.

## 7.6 Domains from Previous Competitions

Tables 9 and 10 show results on the IPC domains from previous years (1998–2006). As these domains do not contain action costs, the cost-sensitive configurations of LAMA are not applicable and LAMA runs with the **FL** configuration. The configurations examined for LAMA are thus **FL** and **F**, both with iterated search and without, where **FL** with iterated search is shown as LAMA. Also given are the results for two IPC-winning systems of previous years, FF and Fast Downward. For both FF and Fast Downward, we ran current versions. In particular Fast Downward has evolved substantially since its 2004 competition version, the original causal graph heuristic having been replaced with the better context-enhanced additive heuristic (Helmert & Geffner, 2008). After correspondence with the authors, the version of Fast Downward used here is the one featuring in recent work by Richter and Helmert (2009).

As Table 9 shows, LAMA performs better than both FF and Fast Downward in terms of the IPC 2008 criterion. This is true even if we turn off landmarks or iterated search in LAMA, but not if we turn off both options simultaneously. When viewing the large difference between the scores of iterated versus non-iterated search in LAMA, note that on these domains no "best known" reference results were used in the score calculation (in contrast to the 2008 tasks, for most of which such reference results were generated manually or with domain-specific solvers by the competition organisers). This means that the planner producing the best solution for a task is awarded the highest-possible score of 1, even though better solutions might exist. This may skew results in favour of the planner that delivers cheaper solutions, i. e., exaggerate the differences between planners.

Table 10 shows that LAMA's edge over Fast Downward is due to higher-quality solutions rather than coverage, as Fast Downward solves more problems. Compared to FF, LAMA has better coverage, with the gap between LAMA and FF being substantially larger than the gap between LAMA and Fast Downward. Note that the **F** and LAMA configurations roughly correspond to the results published as "base" and "heur" in earlier work (Richter et al., 2008). However, subsequent changes to the code to support action costs negatively affect in particular the Philosophers domain, where we observe a significant decrease in coverage. This is also one of the reasons for the difference in coverage between LAMA and the closely related Fast Downward system.

Comparing the various experimental configurations for LAMA, we note that the use of landmarks leads to moderate improvements in both coverage and solution quality. As mentioned above, iterative search significantly improves performance in terms of the IPC 2008 score.

## 8. Conclusion and Outlook

In this article, we have given a detailed account of the LAMA planning system. The system uses two heuristic functions in a multi-heuristic state-space search: a cost-sensitive version of the FF heuristic, and a landmark heuristic guiding the search towards states where many subgoals have already been achieved. Action costs are employed by the heuristic functions to guide the search to cheap goals rather than close goals, and iterative search improves solution quality while time remains.

| Domain | FF | F. Downw. | LAMA | F | FL^first | F^first |
|---|---|---|---|---|---|---|
| **Airport** (50) | 35 | 39 | 35 | 33 | 35 | 33 |
| **Assembly** (30) | 29 | 28 | 30 | 30 | 29 | 29 |
| **Blocks** (35) | 30 | 17 | 33 | 34 | 22 | 17 |
| **Depot** (22) | 20 | 13 | 16 | 15 | 13 | 12 |
| **Driverlog** (20) | 13 | 14 | 19 | 20 | 16 | 15 |
| **Freecell** (80) | 69 | 66 | 73 | 75 | 62 | 65 |
| **Grid** (5) | 4 | 4 | 5 | 5 | 4 | 4 |
| **Gripper** (20) | 20 | 15 | 20 | 20 | 20 | 18 |
| **Logistics 1998** (35) | 35 | 33 | 34 | 33 | 33 | 32 |
| **Logistics 2000** (28) | 28 | 25 | 28 | 28 | 28 | 28 |
| **Miconic** (150) | 150 | 118 | 150 | 143 | 150 | 117 |
| **Miconic Full ADL** (150) | 124 | 95 | 136 | 136 | 107 | 107 |
| **Miconic Simple ADL** (150) | 140 | 105 | 148 | 150 | 117 | 113 |
| **Movie** (30) | 30 | 30 | 30 | 30 | 30 | 30 |
| **MPrime** (35) | 28 | 34 | 35 | 33 | 31 | 29 |
| **Mystery** (30) | 14 | 18 | 19 | 16 | 18 | 14 |
| **Openstacks** (30) | 29 | 29 | 29 | 30 | 29 | 30 |
| **Optical Telegraphs** (48) | 12 | 4 | 2 | 2 | 2 | 2 |
| **Pathways** (30) | 19 | 28 | 28 | 27 | 28 | 27 |
| **Philosophers** (48) | 11 | 48 | 29 | 34 | 29 | 34 |
| **Pipesworld Notank.** (50) | 25 | 31 | 43 | 42 | 26 | 27 |
| **Pipesworld Tank.** (50) | 16 | 28 | 36 | 38 | 27 | 28 |
| **PSR Small** (50) | 41 | 49 | 50 | 50 | 49 | 49 |
| **Rovers** (40) | 38 | 35 | 39 | 39 | 37 | 37 |
| **Satellite** (36) | 35 | 30 | 33 | 31 | 32 | 27 |
| **Schedule** (150) | 99 | 132 | 147 | 139 | 137 | 129 |
| **Storage** (30) | 16 | 16 | 19 | 20 | 16 | 18 |
| **TPP** (30) | 23 | 26 | 30 | 29 | 28 | 27 |
| **Trucks** (30) | 10 | 13 | 13 | 16 | 12 | 15 |
| **Zenotravel** (20) | 19 | 17 | 19 | 20 | 18 | 18 |
| **Total (1512)** | **1162** | **1143** | **1330** | **1318** | **1185** | **1129** |
| **PSR Large** (50) | — | 26 | 28 | 16 | 22 | 14 |
| **PSR Middle** (50) | — | 40 | 50 | 41 | 37 | 35 |

Table 9: Performance scores (rounded to whole numbers) for FF, Fast Downward and LAMA as well as experimental alternative configurations of LAMA (**F**: without landmarks, **FL**^first: without iterated search, **F**^first: without landmarks and without iterated search).

| Domain | FF | F. Downw. | LAMA | F |
|---|---|---|---|---|
| **Airport** (50) | 37 | 40 | 36 | 34 |
| **Assembly** (30) | 30 | 30 | 30 | 30 |
| **Blocks** (35) | 31 | 35 | 35 | 35 |
| **Depot** (22) | 22 | 19 | 17 | 16 |
| **Driverlog** (20) | 15 | 20 | 20 | 20 |
| **Freecell** (80) | 80 | 79 | 79 | 78 |
| **Grid** (5) | 5 | 5 | 5 | 5 |
| **Gripper** (20) | 20 | 20 | 20 | 20 |
| **Logistics 1998** (35) | 35 | 35 | 35 | 35 |
| **Logistics 2000** (28) | 28 | 28 | 28 | 28 |
| **Miconic** (150) | 150 | 150 | 150 | 150 |
| **Miconic Full ADL** (150) | 136 | 139 | 137 | 138 |
| **Miconic Simple ADL** (150) | 150 | 150 | 150 | 150 |
| **Movie** (30) | 30 | 30 | 30 | 30 |
| **MPrime** (35) | 34 | 35 | 35 | 35 |
| **Mystery** (30) | 16 | 19 | 19 | 16 |
| **Openstacks** (30) | 30 | 30 | 30 | 30 |
| **Optical Telegraphs** (48) | 13 | 5 | 2 | 2 |
| **Pathways** (30) | 20 | 29 | 29 | 28 |
| **Philosophers** (48) | 13 | 48 | 29 | 34 |
| **Pipesworld Notank.** (50) | 36 | 43 | 44 | 43 |
| **Pipesworld Tank.** (50) | 21 | 38 | 38 | 40 |
| **PSR Small** (50) | 41 | 50 | 50 | 50 |
| **Rovers** (40) | 40 | 39 | 40 | 40 |
| **Satellite** (36) | 36 | 35 | 34 | 31 |
| **Schedule** (150) | 133 | 150 | 150 | 144 |
| **Storage** (30) | 18 | 18 | 19 | 20 |
| **TPP** (30) | 28 | 30 | 30 | 30 |
| **Trucks** (30) | 11 | 15 | 13 | 16 |
| **Zenotravel** (20) | 20 | 20 | 20 | 20 |
| **Total** (1512) | **1279** | **1384** | **1354** | **1348** |
| **PSR Large** (50) | — | 31 | 29 | 16 |
| **PSR Middle** (50) | — | 50 | 50 | 41 |

Table 10: Coverage (problems solved) for FF, Fast Downward and LAMA as well as the experimental **F** configuration of LAMA without landmarks.

We have conducted an extensive experimental study on the set of benchmark tasks from the last international planning competition, in order to identify how much each of the features of our planner contributes to its performance in the setting of planning with action costs. We discussed overall results and provided plausible explanations for deviating behaviour in some special cases.

The most noticeable outcome of our experiments is that using cost-sensitive heuristics did not produce the desired outcome. In particular, the cost-sensitive FF/add heuristic performs significantly worse than the FF/add heuristic that ignores costs. This is due to the cost-sensitive heuristic solving far fewer tasks while leading to little improvement in solution quality on the tasks that it does solve, especially when using iterated search. When investigating the reasons for this effect, we found that the cost-sensitive FF/add heuristic reacts strongly to bad relaxed plans, i. e., it is in particular in those domains where the relaxed plans computed by the heuristic have low quality that the cost-sensitive heuristic is likely to perform worse than the cost-unaware heuristic. As we showed for the Elevators domain, action costs may also introduce local minima into the search space where without action costs the search space of the FF/add heuristic would have plateaus. Moreover, the increased complexity of planning for a cheaper goal that is potentially further away from the initial state may lead to worse performance.

Landmarks prove to be very helpful in this context, as they mitigate the problems of the cost-sensitive FF/add heuristic. Using landmarks, the coverage of cost-sensitive search is improved to nearly the same level as that of cost-unaware search, while not deteriorating solution quality. Despite the mitigating effect of landmarks, however, LAMA would still have achieved a slightly higher score at IPC 2008 if it had simply ignored costs, rather than using cost-sensitive heuristics. For *cost-unaware* search, we found landmarks to improve coverage and solution quality in the domains from the IPCs 1998–2006. On the domains from IPC 2008, landmarks improved solution quality for the cost-unaware search, but did not further increase the (already very high) coverage.

Iterative search improves results notably for all of our experimental configurations, raising the score of LAMA by a quarter on the IPC 2008 domains. In the Openstacks domain, we could furthermore observe a synergy effect between the iterative search and landmarks. While landmarks usually improve quality, in this domain they lead to bad plans by not accounting for action costs. However, they speed up planning so that the planner evaluates substantially fewer states. Iterative search then effectively improves on the initial bad plans while benefiting from the speed-up provided by the landmarks. In general, we can use landmarks as a means to quickly find good solutions, while using iterative search as a way to improve plan quality over time. Overall, we found that the domains used at IPC 2008 constitute a varied benchmark set that reveals various strengths and weaknesses in our planning system.

Building on the results presented in this article, we identify several directions for future work. Firstly, our results suggest that more research into cost-sensitive heuristics is needed. We would like to conduct a more thorough analysis of the short-comings of the cost-sensitive FF/add heuristic, to answer the question whether and how they might be overcome. Keyder and Geffner (2009) propose a method for extracting better relaxed plans from the best supports computed by the cost-sensitive FF/add heuristic, resulting in improved coverage. However, the large ledge of the cost-unaware heuristic in our experiments suggests that the cost-unaware FF/add heuristic is still better than the improved cost-sensitive heuristic by Keyder and Geffner. It would be interesting to examine to what degree the problems we experienced with the FF/add heuristic extend to other delete-relaxation heuristics, and whether heuristics not based on the delete relaxation could be more effectively adapted to action costs. In addition, future work could explore the benefit of combin-ing traditional distance estimators and cost-sensitive heuristics in more sophisticated ways than the mechanism currently used in LAMA (see the discussion in Section 3.3.2).

Secondly, we believe it to be useful for future research to improve the definition of reasonable orderings, eliminating the problems of the definition by Hoffmann et al. mentioned in Section 4.1.

Thirdly, we would like to extend the use of landmarks in our system in several ways. For one, our current approach does not take into account whether the same landmark must be achieved several times. Supporting such multiple occurrences of landmarks would be beneficial in the Openstacks domain, for example, as it could help to minimise the creation of stacks by accounting for their costs. While methods exist for detecting the multiplicity of landmarks (Porteous & Cresswell, 2002; Zhu & Givan, 2003), it will be crucial to develop techniques for deriving orderings between the individual occurrences of such landmarks. Furthermore, we would like to extend LAMA to support more complex landmarks like conjunctions or other simple formulas. In addition to representing and using such landmarks in the landmark heuristic this involves the development of new methods for detecting them along with their corresponding orderings.

## Acknowledgments

The authors thank Malte Helmert, Charles Gretton, Sylvie Thiebaux and Patrik Haslum as well as the anonymous reviewers for helpful feedback on earlier drafts of this paper.

The computing resources for our experiments were graciously provided by Pompeu Fabra University. We thank Héctor Palacios for his support in conducting the experiments.

NICTA is funded by the Australian Government, as represented by the Department of Broadband, Communications and the Digital Economy, and the Australian Research Council, through the ICT Centre of Excellence program.

This work was partially supported by Deutsche Forschungsgemeinschaft as part of the Transregional Collaborative Research Center *SFB/TR 8 Spatial Cognition*, project R4-[LogoSpace].